\documentclass{article} 
\usepackage{iclr2025_conference,times}

\usepackage{amsmath,amsfonts,bm}

\def\eqref#1{equation~\ref{#1}}

\def\1{\bm{1}}

\DeclareMathAlphabet{\mathsfit}{\encodingdefault}{\sfdefault}{m}{sl}
\SetMathAlphabet{\mathsfit}{bold}{\encodingdefault}{\sfdefault}{bx}{n}

\usepackage{graphicx}
\usepackage{xspace}
\usepackage{xcolor}
\usepackage{wrapfig}
\usepackage{pifont}

\newcommand{\framework}{\textbf{DreamVideo-2}\xspace}
\newcommand{\frameworkplain}{DreamVideo-2\xspace}

\newcommand{\tabincell}[2]{\begin{tabular}{@{}#1@{}}#2\end{tabular}}

\newlength\savewidth\newcommand\shline{\noalign{\global\savewidth\arrayrulewidth\global\arrayrulewidth1.25pt}\hline\noalign{\global\arrayrulewidth\savewidth}}

\newcommand{\ie}{\textit{i.e.}}
\newcommand{\eg}{\textit{e.g.}}

\definecolor{lightgreen}{rgb}{0.56, 0.93, 0.56}

\definecolor{mycolor_blue}{HTML}{E7EFFA}
\definecolor{mycolor_green}{HTML}{E6F8E0}
\definecolor{mycolor_gray}{HTML}{ECECEC}
\definecolor{pearDark}{HTML}{2980B9}

\definecolor{darkred}{rgb}{0.7,0.1,0.1}
\definecolor{darkgreen}{rgb}{0.1,0.6,0.1}
\newcommand{\cmark}{\color{darkgreen}{\ding{51}}}
\newcommand{\xmark}{\color{darkred}{\ding{55}}}

\usepackage{hyperref}
\usepackage{url}

\title{\frameworkplain: Zero-Shot Subject-Driven Video Customization with Precise Motion Control}

\author{%
 Yujie Wei$^{1}$, Shiwei Zhang$^2$\thanks{Project Leader\quad  $^\dagger$Corresponding Author} , Hangjie Yuan$^2$, Xiang Wang$^2$, Haonan Qiu$^3$, Rui Zhao$^2$, \\ \textbf{Yutong Feng$^2$, Feng Liu$^2$, Zhizhong Huang$^4$, Jiaxin Ye$^1$, Yingya Zhang$^2$, Hongming Shan}$^{1\dagger}$
 \\\\
 $^1$Fudan University\qquad$^2$Alibaba Group
 \\
 $^3$Nanyang Technological University\qquad
 $^4$Michigan State University 
 \\ \vspace{-0.6em} \\ 
 Project page: \url{https://dreamvideo2.github.io}
}

\iclrfinalcopy 
\begin{document}

\maketitle

\begin{figure}[h!]
\vspace{-2.1em}
\centering
\includegraphics[width=1\linewidth]{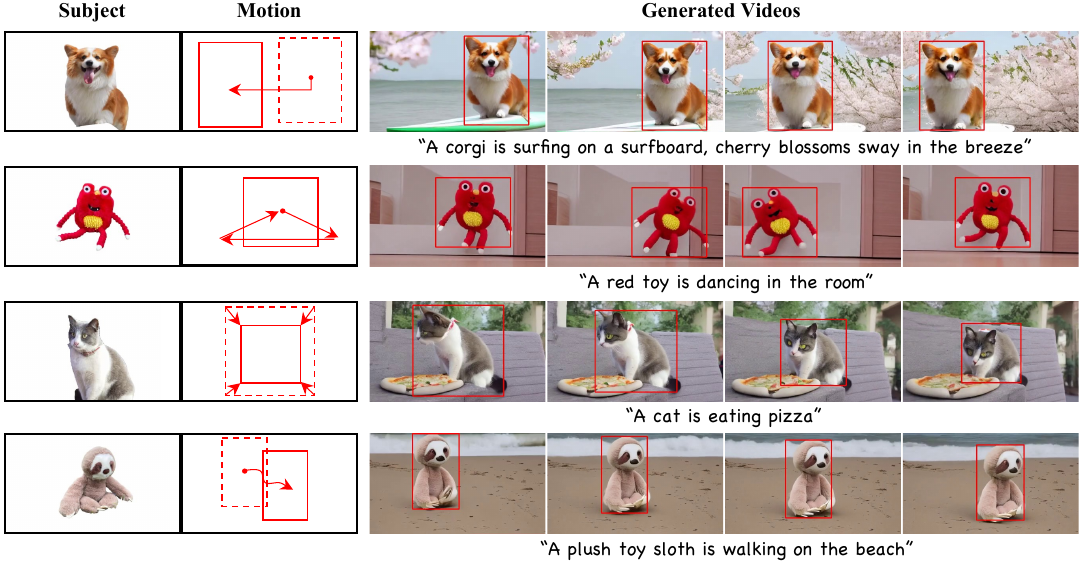}
\vspace{-2em}
\caption{\textbf{Customized video generation results of \framework}. Our method precisely generates customized subjects at specified positions
\textbf{without fine-tuning at inference time}. 
}
\vspace{-0.1em}
\label{fig:teaser}
\end{figure}

\begin{abstract}
Recent advances in customized video generation have enabled users to create videos tailored to both specific subjects and motion trajectories.
However, existing methods often require complicated test-time fine-tuning and struggle with balancing subject learning and motion control, limiting their real-world applications.
In this paper, we present \framework, a zero-shot video customization framework capable of generating videos with a specific subject and motion trajectory, guided by a single image and a bounding box sequence, respectively, and without the need for test-time fine-tuning.
Specifically, we introduce reference attention, which leverages the model's inherent capabilities for subject learning, and devise a mask-guided motion module to achieve precise motion control by fully utilizing the robust motion signal of box masks derived from bounding boxes.
While these two components achieve their intended functions, we empirically observe that motion control tends to dominate over subject learning.
To address this, we propose two key designs:
\textbf{1}) the masked reference attention, which integrates a blended latent mask modeling scheme into reference attention to enhance subject representations at the desired positions, and 
\textbf{2}) a reweighted diffusion loss, which differentiates the contributions of regions inside and outside the bounding boxes to ensure a balance between subject and motion control.
Extensive experimental results on a newly curated dataset demonstrate that \frameworkplain
outperforms state-of-the-art methods in both subject customization and motion control.
The dataset, code, and models will be made publicly available. 
\end{abstract}
\section{Introduction}
Customized video generation~\citep{dreamix, motionDirector, wei2024dreamvideo, chen2023videodreamer} has made significant strides, largely driven by the remarkable advances in pre-trained text-to-video generation models~\citep{VDM, LVDM, modelScope, wang2023lavie, videocrafter1, hong2022cogvideo, yang2024cogvideox}.
These innovations enable users to create videos with specific subjects and precise motion trajectories~\citep{wu2024customcrafter, direct_a_video, wang2024motionctrl}, thereby broadening the scope of real-world applications for video generation.

Pioneering research efforts have explored customized video generation~\citep{chen2023videodreamer, jeong2024vmc, jiang2024videobooth, wei2024dreamvideo}, but they encounter significant limitations in: 
(1) the lack of comprehensive control over subjects and motions in a zero-shot manner, and 
(2) the conflict between subject learning and motion control.
For instance, VideoBooth~\citep{jiang2024videobooth} employs a tuning-free framework to inject subject embeddings from image prompts for subject customization, but it fails to control motion dynamics,  leading to generated videos with minimal or absent motion.
In contrast,
some fine-tuning-based approaches attempt to control subject and motion simultaneously.
For example, DreamVideo~\citep{wei2024dreamvideo} trains two adapters separately and combines them during inference,
while MotionBooth~\citep{wu2024motionbooth} trains a customized model and manipulates attention maps to control motion during inference.
However, an empirical training-inference gap persists, preventing these methods from achieving a balance between subject and motion learning.
Therefore, \emph{simultaneously enhancing and balancing subject learning and motion control in a zero-shot manner} holds great potential for practical video customization.

To that end, we propose an innovative zero-shot video customization framework,  \framework, which can generate videos with a specified subject and motion trajectory, derived from a \emph{single} image and a bounding box sequence, respectively, as illustrated in Fig.~\ref{fig:teaser}. 
\frameworkplain concurrently learns subject appearance and motion during training, allowing for harmonious subject and motion control without additional fine-tuning or manipulation during inference.
To effectively inject detailed appearance information from a subject image, we introduce reference attention that leverages multi-scale features extracted from the original video diffusion model.
For motion control, we devise a mask-guided motion module comprised of a spatiotemporal encoder and a spatial ControlNet~\citep{controlnet}, which adopts binary box masks derived from the bounding boxes as the robust motion control signal, significantly improving control precision.

While these two components can achieve their intended functions of subject and motion control, systematic experiments empirically reveal that motion control tends to dominate over subject learning, partially due to the simpler objective of generating subjects at specified positions, which compromises subject preservation quality.
To mitigate this issue, we aim to strengthen the learning of subjects with two new technical contributions:
\textbf{1}) the masked reference attention, which  introduces a blended latent mask modeling scheme into our reference attention to enhance subject identity representations at desired positions by leveraging box masks; and 
\textbf{2}) a reweighted diffusion loss function, which differentiates the contributions of regions inside and outside the bounding boxes to ensure a balance between subject and motion control.

To facilitate the zero-shot video customization task, we curate a new single-subject video dataset with comprehensive annotations, comprising the caption and each frame's subject mask and bounding box.
This dataset is not only larger but also considerably more diverse than previous video customization datasets.
Extensive experimental results on this dataset demonstrate that \frameworkplain
outperforms state-of-the-art methods in both customization and control capabilities.

\noindent\textbf{Contributions.}\quad 
The contributions 
of this work 
can be summarized as follows.
\textbf{1)} We propose \frameworkplain, the first tuning-free framework 
for zero-shot subject-driven video customization with precise motion trajectory control, achieved through the devised reference attention and the mask-guided motion module that uses binary box masks as motion control signals.
\textbf{2)} We identify the problem of motion control dominance in \frameworkplain, and address it by enhancing reference attention with blended masks (\textit{i.e.}, masked reference attention) and designing a reweighted diffusion loss, effectively balancing subject learning and motion control.
\textbf{3)} We curate a large, comprehensive, and diverse video dataset to support the zero-shot video customization task. 
Extensive experimental results demonstrate the superiority of \frameworkplain over the existing state-of-the-art video customization methods.
\section{Related Work}
\label{sec:related_work}
\noindent\textbf{Text-to-video diffusion models.}\quad
Diffusion models have made a significant breakthrough in the generation of highly realistic samples from diverse prompts~\citep{DDPM, stableDiffusion, podell2023sdxl, mou2024t2i, li2024unihda, wang2024fldm, zhao2024evolvedirector, li2023few}.
Recent advancements in text-to-video generation have expanded upon these models by incorporating temporal dynamics, enabling the production of high-quality and diverse video content~\citep{gen1, latent_shift, show1, i2vgen_xl, higen, wang2023videolcm, tft2v, make-a-video, imagenVideo, magicVideo, wang2023lavie, yuan2024instructvideo, ma2024latte, gupta2023photorealistic, bar2024lumiere, videoComposer, tan2024animateX}. 
VDM~\citep{VDM} first introduces diffusion models into video generation by modeling the video distribution in pixel space.
VLDM~\citep{align-your-latents} optimizes the diffusion process in the latent space to mitigate computational demands. 
ModelScopeT2V~\citep{modelScope} and VideoCrafter~\citep{videocrafter1, chen2024videocrafter2} incorporate spatiotemporal blocks for text-to-video generation.
AnimateDiff~\citep{animatediff} trains a motion module appended to the pre-trained text-to-image models.
SVD~\citep{svd} enhances the scalability of the latent video diffusion model.
VideoPoet~\citep{kondratyuk2023videopoet} investigates autoregressive video generation.
Sora~\citep{sora} significantly improves the quality and stability of video generation.
These advanced video generative models pave the way for customized video generation.

\noindent\textbf{Customized generation.}\quad
Customized image generation has garnered growing attention since it accommodates user preferences~\citep{disenbooth, han2023svdiff, SuTI, wei2023elite, shi2024instantbooth, li2024stylegan, ruiz2024hyperdreambooth, hua2023dreamtuner, han2024face_adapter, mix_of_show, cones, fastcomposer, customDiffusion, cones2, anydoor}. 
The representative works are Textual Inversion~\citep{textInversion} and DreamBooth~\citep{dreambooth}, where Textual Inversion optimizes text embeddings and DreamBooth fine-tunes an image diffusion model.
Building upon these methods, many works explore customized video generation using a few subject or facial images~\citep{dreamix, chefer2024still_moving, ma2024magicme, he2024id_animator}.
Furthermore, several works study the more challenging multi-subject video customization task~\citep{chen2023videodreamer, wang2024customvideo, chen2024disenstudio}.
Considering that spatial content and temporal dynamics are two indispensable components of videos, DreamVideo~\citep{wei2024dreamvideo} customizes both subject and motion by training two adapters and combining them at inference time, while 
MotionBooth~\citep{wu2024motionbooth} fully fine-tunes a video diffusion model to learn subjects during training and edits the attention maps to control motion during inference.
However, 
both methods require complicated test-time fine-tuning and struggle with balancing subject and motion control due to an empirical training-inference gap.
In contrast, our \frameworkplain generates videos with harmonious subject and motion control in a tuning-free manner.

\noindent\textbf{Motion control in video generation.}\quad
Recent advancements in controllable video generation primarily focus on enhancing motion dynamics through additional control signals.
Many motion customization methods learn motion patterns from intuitive reference videos~\citep{motionDirector, jeong2024vmc, customize_a_video, DMT, motionInversion, lamp}, 
but they often require complicated fine-tuning for each motion at inference time. 
To circumvent the need for fine-tuning, some training-free methods manipulate attention map values through bounding boxes to control the object movements~\citep{jain2024peekaboo, direct_a_video, ma2023trailblazer, motion_zero, qiu2024freetraj}, 
However, these methods fail to achieve precise motion control, resulting in inconsistent frames. 
In contrast, several works use trajectories or coordinates as additional conditions to train a motion control module~\citep{yin2023dragnuwa, wang2024motionctrl, wang2024boximator, imageConductor}. 
Nonetheless, they tend to achieve general motion control but fail to incorporate user-specified object appearances, which may limit their practical applicability.
In this work, we propose masked reference attention and devise a mask-guided motion module to control the subject and motion simultaneously, effectively mitigating the control conflict using a devised reweighted diffusion loss. 
\section{Preliminary}
\label{sec:background}
\textbf{Video diffusion models.}\quad
Video diffusion models (VDMs)~\citep{VDM} aim to generate video data using diffusion processes~\citep{DDPM}.
Most VDMs~\citep{align-your-latents, modelScope, videoComposer} perform the diffusion processes in a latent space using a VAE~\citep{vae} encoder $\mathcal{E}$ to map a video $\bm{x}_0 \in \mathbb{R}^{F \times H \times W \times 3}$ into its latent code $\bm{z}_0 = \mathcal{E}(\bm{x}_0), \bm{z}_0 \in \mathbb{R}^{F \times h \times w \times 4}$, and a decoder $\mathcal{D}$ to reconstruct the video $\hat{\bm{x}}_0 = \mathcal{D}(\bm{z}_0)$. The forward process gradually adds noise to the latent code $\bm{z}_0$ according to a predetermined schedule ${\{\beta_t\}}_{t=1}^{T}$ with $T$ steps:
$
    \bm{z}_t = \sqrt{\bar{\alpha}_t} \bm{z} + \sqrt{1 - \bar{\alpha}_t} \epsilon, 
$
where $ \bar{\alpha}_t = \prod_{s=1}^t \alpha_s$, $\alpha_t = 1 - \beta_t$, and $\epsilon  \in \mathcal{N}(0,1)$ is random noise from a Gaussian distribution.

The reverse process adopts a network $\epsilon_{\theta}$ to predict the added noise $\epsilon$ at each timestep $t$ based on an additional condition $\bm{c}$. The training objective can be simplified as a reconstruction loss:
\begin{equation}
    \mathcal{L}(\theta) = \mathbb{E}_{\bm{z}, \epsilon, \bm{c}, t} \left[\left\| \epsilon - \epsilon_{\theta}(\bm{z}_t, \bm{c}, t) \right\|_{2}^{2}\right].
\end{equation}

\textbf{Attention mechanism in VDMs.}\quad
In most text-to-video VDMs, self-attention serves to capture contextual features, while cross-attention facilitates the integration of additional conditions, such as textual features $\bm{c}_\text{txt}$.
Given the features $\mathbf{Z}$ from the latent code, the standard formulation of the attention mechanism can be expressed as:
\begin{equation}
\mathbf{Z^{\prime}} = \operatorname{Attention}(\mathbf{Q}, \mathbf{K}, \mathbf{V})
= \operatorname{Softmax}\left(\frac{\mathbf{Q K}^{\top}}{\sqrt{d}}\right) \mathbf{V},
\label{eq:attention}
\end{equation}
where
$\mathbf{Z^{\prime}}$ is the output attention features.
$\mathbf{Q}$, $\mathbf{K}$, and $\mathbf{V}$ are the query, key, and value matrices, respectively. 
For self-attention, $\mathbf{Q} = \mathbf{Z} \mathbf{W}_Q$, $\mathbf{K} = \mathbf{Z} \mathbf{W}_K$, $\mathbf{V} = \mathbf{Z} \mathbf{W}_V$ , and for cross-attention, $\mathbf{Q} = \mathbf{Z} \mathbf{W}_Q$, $\mathbf{K} = \bm{c} \mathbf{W}_K$, $\mathbf{V} = \bm{c} \mathbf{W}_V$. 
Here, $\mathbf{W}_Q$, $\mathbf{W}_K$, $\mathbf{W}_V$ are the corresponding projection matrices. 
$d$ is the dimension of key features.

\begin{figure}[t]
  \centering
  \includegraphics[width=1.0\linewidth]{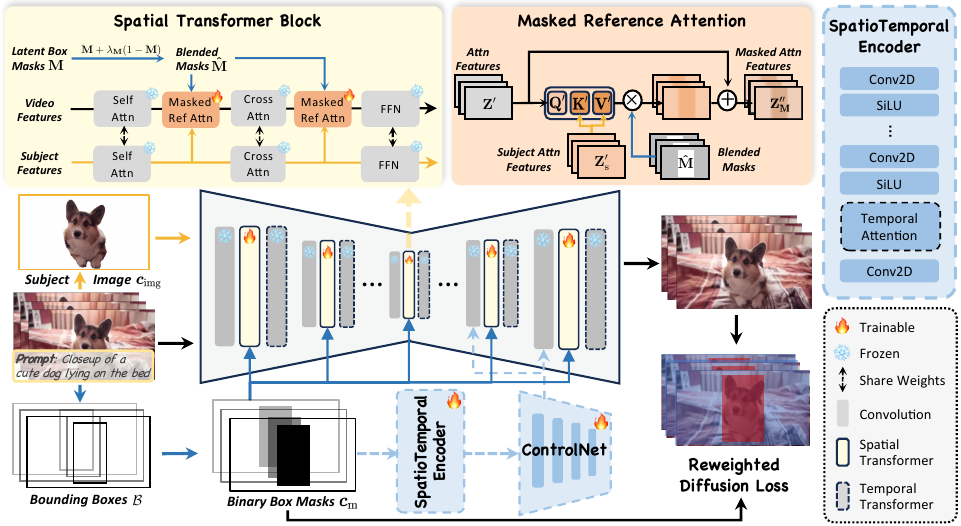}
  \caption{\textbf{Overall framework of \framework}. During training, a random video frame is segmented to obtain the subject image with a blank background.
  The bounding boxes extracted from the training video are converted into binary box masks. 
  Then, the subject image is treated as a single-frame video and processed in parallel with the video by masked reference attention that incorporates blended masks to learn the subject appearance.
  Meanwhile, box masks are fed into a motion module that includes a spatiotemporal encoder and a ControlNet for motion control. 
  Both the masked reference attention and motion module are trained using a reweighted diffusion loss.}
  \label{fig:framework}
\end{figure}

\section{Methodology}
\label{sec:method}
Given a single subject image that defines the subject's appearance and a bounding box sequence that delineates the motion trajectory,
our \framework aims to generate videos featuring specified subjects and motion trajectories without fine-tuning or manipulation at inference time,  as illustrated in Fig.~\ref{fig:framework}.
To learn the subject appearance, we leverage the model’s inherent capabilities and introduce reference attention in Sec.~\ref{sec:method_id}. For motion control, we propose using box masks as the motion control signal and devise a mask-guided motion module in Sec.~\ref{sec:method_motion}.
Furthermore, to balance subject learning and motion control, we enhance reference attention with blended masks (\ie, masked reference attention) and design a reweighted diffusion loss in Sec.~\ref{sec:method_balance}.
Finally, we detail the training, inference, and dataset construction processes in Sec.~\ref{sec:method_train_infer}.

\subsection{Subject Learning via Reference Attention}
\label{sec:method_id}
For subject learning, we focus on using a single image to capture the appearance details, which is challenging but facilitates real-world applications.
Given a single input image, we first segment it to obtain the subject image $\mathbf{c}_{\text{img}}$ with a blank background, effectively preserving distinct identity features while minimizing background interference~\citep{chen2024anydoor, jiang2024videobooth}.

To capture the intricate details of the subject's appearance, previous works usually employ an extra image encoder (\eg, CLIP~\citep{ye2023ip_adapter, jiang2024videobooth}, ControlNet-like encoder~\citep{anydoor}, ReferenceNet~\citep{hu2024animateanyone}) to extract image features.
However, incorporating additional networks tends to escalate both parameter counts and training costs.
In this work, we identify that the video diffusion model itself is capable of extracting appearance features, thus improving training efficiency without requiring auxiliary modules.

To that end, we introduce reference attention, which leverages the model’s inherent capabilities to extract multi-scale subject features. 
Specifically, we treat the subject image as a single-frame video and input it into the original video diffusion model to obtain subject attention features $\mathbf{Z}^{\prime}_{\text{s}}$, which is the output of self-attention or cross-attention according to Eq.~(\ref{eq:attention}).
Our reference attention infuses the subject attention features into video attention features $\mathbf{Z}^{\prime}$ by implementing a residual cross-attention:
\begin{equation}
\mathbf{Z^{\prime\prime}} = \mathbf{Z^{\prime}} + \operatorname{Attention}(\mathbf{Q^{\prime}}, \mathbf{K^{\prime}}, \mathbf{V^{\prime}}),
\label{eq:ref_attention}
\end{equation}
where $\mathbf{Q}^{\prime} = \mathbf{Z}^{\prime} \mathbf{W}_Q^{\prime}$, $\mathbf{K}^{\prime} = \mathbf{Z}^{\prime}_{\text{s}} \mathbf{W}_K^{\prime}$, $\mathbf{V}^{\prime} = \mathbf{Z}^{\prime}_{\text{s}} \mathbf{W}_V^{\prime}$.
$\mathbf{W}_Q^{\prime}$, $\mathbf{W}_K^{\prime}$, and $\mathbf{W}_V^{\prime}$ are the projection matrices of reference attention and are initialized randomly. 
In addition, we initialize the weights of the output linear layer in reference attention with zeros to protect the pre-trained model from being damaged at the beginning of training~\citep{controlnet, wei2024dreamvideo}.

\subsection{Motion Control via Mask-Guided Motion Module}
\label{sec:method_motion}
To facilitate motion control, we utilize bounding boxes as user inputs to delineate subject trajectories, offering both flexibility and convenience.
We define an input sequence of bounding boxes as $\mathcal{B} = [\mathcal{B}_1, \mathcal{B}_2, \ldots, \mathcal{B}_F]$, where each box $\mathcal{B}_i$
includes coordinates of its top-left and bottom-right corners.
Then, we convert these bounding boxes into a binary box mask sequence $ \mathcal{M} = [\mathcal{M}_1, \mathcal{M}_2, \ldots, \mathcal{M}_F]$, where each mask $\mathcal{M}_i \in \mathbb{R}^{H \times W}$ has pixel values of 1 for the foreground subject and 0 for the background.

The final motion control signal is represented as $\bm{c}_{\text{m}} = 1 - \mathcal{M} $ to align with the subject image containing a blank background.
Compared to directly using trajectories for training in previous work~\citep{wang2024motionctrl}, the box masks provide enhanced control signals and constrain subjects within the bounding box, improving training efficiency and motion control precision.

To capture motion information from the box mask sequence, we devise a mask-guided motion module, which employs a spatiotemporal encoder and a spatial ControlNet~\citep{controlnet}, as depicted in Fig.~\ref{fig:framework}.
While previous research~\citep{guo2023sparsectrl} demonstrates the efficacy of a 3D ControlNet for extracting control information from sequential inputs, its high training costs present potential drawbacks in practical applications.
Given the straightforward temporal relationships in the box mask sequence, we establish that a lightweight spatiotemporal encoder is adequate for extracting the necessary temporal information. 
Thus, we only employ a spatial ControlNet appended to this encoder to further enhance control precision.
The spatiotemporal encoder consists of repeated 2D convolutions and non-linear layers, followed by two temporal attention layers and an output convolutional layer, as shown in the right side of Fig.~\ref{fig:framework}.
In addition, the spatial ControlNet extracts multi-scale features and adds them to the input of convolutional layers of the VDM's decoder blocks.

\subsection{Balancing Subject Learning and Motion Control}
\label{sec:method_balance}
\begin{wrapfigure}{r}{0.5\textwidth}
\centering
\vspace{-4mm}
\includegraphics[width=1\linewidth]{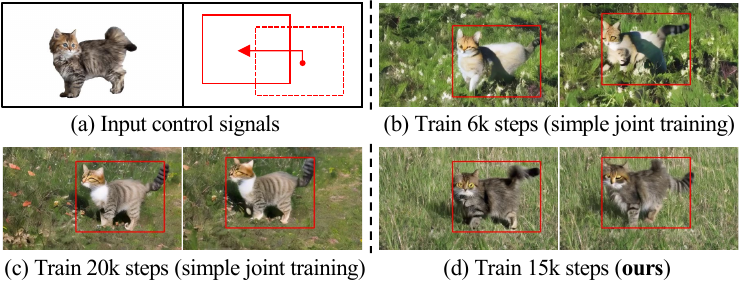}
\vspace{-2em}
\caption{\textbf{Illustration of motion control domination in \framework}. As seen in (b) and (c), motion control tends to dominate over subject learning during training, causing the degradation of subject identity. 
In (d), our method ensures a balance between subject and motion control.
}
\label{fig:motion_dominate}
\vspace{-1em}
\end{wrapfigure}

While the above two components achieve their intended functions, we empirically observe that motion control tends to dominate over subject learning, which compromises identity preservation quality.
As shown in Fig.~\ref{fig:motion_dominate}(b), the model learns motion control using a few steps, partially due to the simpler objective of generating subjects at specified positions. In Fig.~\ref{fig:motion_dominate}(c), 
joint training of the reference attention and motion module retains the dominance of motion control, even with extended training steps, resulting in corrupted subject identity.
In contrast, as shown in Fig.~\ref{fig:motion_dominate}(d), our method effectively balances subject learning and motion control by proposing the following two key designs.

\textbf{Masked reference attention.}\quad
To enhance the subject identity representations at desired positions, we introduce blended latent mask modeling into our reference attention through binary box masks.
Specifically, we resize the binary box masks $\mathcal{M}$ into latent box masks $\mathbf{M} = [\mathbf{M}_1, \mathbf{M}_2, \ldots, \mathbf{M}_F | \mathbf{M}_i \in \mathbb{R}^{h \times w}]$ to match the size of attention features across different layers. 

Then, we assign a relatively lower weight to the background (\ie, regions outside the bounding boxes) in $\mathbf{M}$
to obtain blended masks $\mathbf{\hat{M}}$, forcing the model to focus more on the subject and less on the background at the feature level:
\begin{equation}
\mathbf{\hat{M}} = \mathbf{M} + \lambda_{\mathbf{M}} (1 - \mathbf{M}),
\end{equation}
where $\lambda_{\mathbf{M}}$ is the weight of background in mask.
Compared to using binary masks $\mathbf{M}$, which ignore background information, blended masks $\mathbf{\hat{M}}$ can enhance the subject representations at desired positions while mitigating the background distortion.
Finally, our masked reference attention can be formulated as:
\begin{equation}
\mathbf{Z}^{\prime\prime}_{\mathbf{M}}= \mathbf{Z}^{\prime} + \mathbf{\hat{M}} \cdot \operatorname{Attention}(\mathbf{Q}^{\prime}, \mathbf{K}^{\prime}, \mathbf{V}^{\prime}),
\end{equation}
where
$\cdot$ denotes the element-wise multiplication operation.
For subject learning, we freeze all original UNet parameters and only train the masked reference attentions, which are appended to both self-attention and cross-attention within each spatial transformer block, as shown in Fig.~\ref{fig:framework}.

\textbf{Reweighted diffusion loss.}\quad
To balance subject learning and motion control, we further propose a reweighted diffusion loss that differentiates the contributions of regions inside and outside the bounding boxes to the standard diffusion loss.
Specifically, we amplify the contributions within bounding boxes to enhance subject learning while preserving the original diffusion loss for regions outside these boxes.
Our designed reweighted diffusion loss can be defined as:
\begin{equation}
    \mathcal{L}(\theta) = \mathbb{E}_{\bm{z}, \epsilon, \bm{c}, t} 
    \left[
    \Big( \underbrace{\lambda_{\mathcal{L}}\mathbf{M}\vphantom{(1 - \mathbf{M})}}_{\text {inside}} 
    + 
    \underbrace{(1 - \mathbf{M})}_{\text {outside}} \Big)
    \cdot
    \Big\| 
    \epsilon - \epsilon_{\theta}(\bm{z}_t, \bm{c}_\text{txt}, 
    \bm{c}_\text{img}, \bm{c}_\text{m},
    t)
    \Big\|_{2}^{2}
    \right],
\label{eq:box_mask_loss}
\end{equation}
where $\lambda_{\mathcal{L}} > 1$ is the loss weight to adjust the subject identity enhancement.

\subsection{Training, Inference, and Dataset Construction}
\label{sec:method_train_infer}
\textbf{Training.}\quad
We randomly select a frame from the training video and segment it to obtain the subject image with a blank background, which alleviates overfitting compared to using the first frame as in~\citep{jiang2024videobooth}.
We also extract the subject's bounding boxes from all frames of the training video and convert them into box masks as the motion control signal.
During training, we freeze the original 3D UNet parameters and jointly train the newly added masked reference attention, spatiotemporal encoder, and  ControlNet according to Eq.~(\ref{eq:box_mask_loss}).

\textbf{Inference.}\quad 
Our \frameworkplain is tuning-free and does not require attention map manipulations during inference.
Users only need to provide a subject image and a bounding box sequence to flexibly generate customized videos featuring the specified subject and motion trajectory. 
The bounding boxes can be derived from various types of signals, including boxes of the first and last frames, a bounding box of the first frame accompanied by a motion trajectory, or a reference video.
These signals are then converted into binary box masks for input.

\textbf{Dataset Construction.}\quad
To facilitate the zero-shot video customization task with subject and motion control,
we curate a single-subject video dataset containing both video masks and bounding boxes from the WebVid-10M~\citep{webvid10m} dataset and our internal data. 
Annotations are generated using the Grounding DINO~\citep{groundingdino}, SAM~\citep{sam}, and DEVA~\citep{cheng2023tracking} models.
The comparison of our dataset and previous datasets is presented in Tab.~\ref{tab:app_dataset_compare}.
Currently, we have processed 230,160 videos for training, and more details are in Appendix~\ref{app:dataset_construction}.

\begin{table}[ht]
    \centering
    \resizebox{\columnwidth}{!}{
    \begin{tabular}{lcccccc} 
         & 
         \tabincell{c}{\textbf{Number of}\\\textbf{Videos}}
        & 
        \tabincell{c}{\textbf{Number of}\\\textbf{Object Classes}}
        &
        \textbf{Caption}& 
        \tabincell{c}{\textbf{Mask of}\\\textbf{All Frames}}
        & 
        \tabincell{c}{\textbf{Box of}\\\textbf{All Frames}}
        \\ 
        \shline
        \noalign{\smallskip}
        WebVid-10M~\citep{webvid10m} & $
        \sim$10M & - & \cmark & \xmark & \xmark \\
        UCF-101~\citep{ucf101} & 13,320 & - & \xmark & \xmark & \xmark \\
        DAVIS~\citep{davis} & 50 & 50 & \xmark & \cmark & \cmark \\
        GOT-10k~\citep{GOT10k} & 9,695 & 563  & \xmark & \xmark & \cmark \\
        VideoBooth Dataset~\citep{jiang2024videobooth} & 48,724 & 9 & \cmark & \xmark & \xmark \\
         \hline
         \noalign{\smallskip}
         \textbf{\framework Dataset} & \textbf{230,160} & \textbf{2,538} & \cmark & \cmark & \cmark \\ 
    \end{tabular}
    }
    \caption{\textbf{Comparsion of our dataset with related video datasets.} Our dataset contains comprehensive annotations, and is larger and more diverse than previous video customization datasets.
    }
    \label{tab:app_dataset_compare}
\end{table}
\section{Experiment}
\label{sec:exp}
\subsection{Experimental Setup}
\label{sec:exp_setup}
\textbf{Datasets.}\quad
We train \frameworkplain on our curated video dataset and evaluate it through a collected test set containing 50 subjects and 36 bounding boxes.
The subject images are sourced from previous papers~\citep{dreambooth, customDiffusion} and the Internet, while bounding boxes are obtained from the videos in DAVIS dataset~\citep{davis} and the boxes used in FreeTraj~\citep{qiu2024freetraj}.
Additionally, we design 60 textual prompts for validation.

\textbf{Implementation details.}\quad
We jointly train all modules using the AdamW~\citep{adamw} optimizer with a learning rate of 1e-4. The weight decay is set to 0, and the training iteration is 30,000. 
We set blended mask weight $\lambda_{\mathbf{M}}$ to 0.75 and reweighted diffusion loss weight $\lambda_{\mathcal{L}}$ to 2 for training. 
The spatial resolution of the videos is 448$\times$256, and the number of video frames $F$ is 16. We set the total batch size to 144, and
adopt ModelScopeT2V~\citep{modelScope} as the base model.
During inference, we employ 50-step DDIM~\citep{DDIM} and classifier-free guidance~\citep{ho2022classifier_free_guide} with guidance scale 9.0 to generate 8-fps videos.

\textbf{Baselines.}\quad
We compare our method with DreamVideo~\citep{wei2024dreamvideo} and MotionBooth~\citep{wu2024motionbooth} for both subject customization and motion control.
We also compare with DreamVideo and VideoBooth~\citep{jiang2024videobooth} for independent subject customization, while Peekaboo~\citep{jain2024peekaboo}, Direct-a-Video~\citep{direct_a_video}, and MotionCtrl~\citep{wang2024motionctrl} for motion trajectory control.
More implementation details of all methods are provided in Appendix~\ref{app:exp_details}.

\textbf{Evaluation metrics.}\quad
We evaluate our method using 9 metrics, focusing on
three aspects: overall consistency, 
subject fidelity, and motion control precision.
\textbf{1)} 
For overall consistency, we employ 
CLIP image-text similarity (CLIP-T), Temporal Consistency (T. Cons.)~\citep{gen1}, and Dynamic Degree (DD)~\citep{huang2024vbench} metrics.
DD uses optical flow to measure motion dynamics.
\textbf{2)}
For subject fidelity, we introduce four metrics: CLIP image similarity (CLIP-I), DINO image similarity (DINO-I), region CLIP-I (R-CLIP), and region DINO-I (R-DINO) metrics~\citep{dreambooth, wei2024dreamvideo, wu2024motionbooth}.
R-CLIP and R-DINO compute the similarities between the subject image and frame regions defined by bounding boxes, following~\citep{wu2024motionbooth}.
\textbf{3)}
For motion control precision, we use the Mean Intersection of Union (mIoU) and Centroid Distance (CD) metrics~\citep{qiu2024freetraj}.
CD computes the normalized distance between the centroid of the generated subject and target bounding boxes.
We use Grounding-DINO~\citep{groundingdino} to predict the bounding boxes of generated videos.
More details of metrics are reported in Appendix~\ref{app:exp_details}.

\subsection{Main Results}
\begin{figure}[t]
  \centering
  \includegraphics[width=1.0\linewidth]{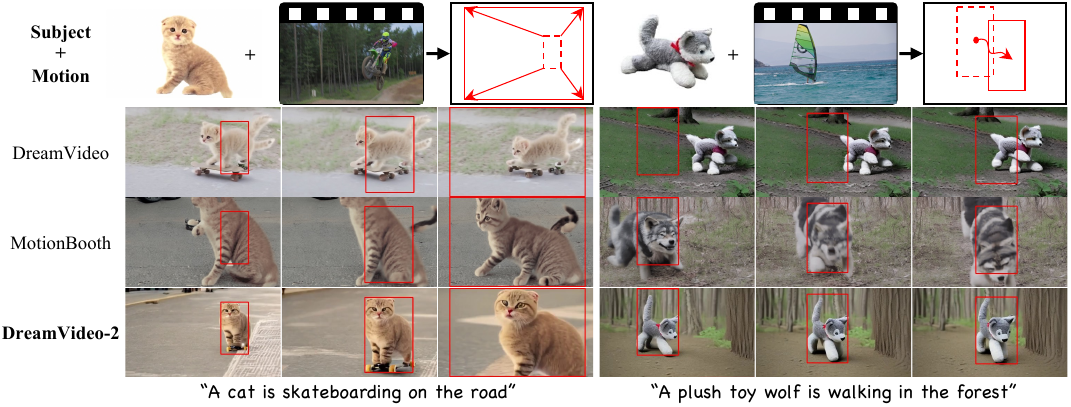}
  \caption{\textbf{Qualitative comparison of joint subject customization and motion control}. \frameworkplain generates videos with customized subjects and precise motion trajectory control, while other methods suffer from control conflicts, especially when trained on a single image.}
  \label{fig:compare_id_motion}
\end{figure}

\textbf{Joint subject customization and motion control.}\quad
We conduct a qualitative comparison between our method and baselines for generating videos featuring both specified subjects and motion trajectories, as depicted in Fig.~\ref{fig:compare_id_motion}.
We observe that DreamVideo and MotionBooth struggle with balancing subject preservation and motion control, especially when trained on a single subject image.
We argue that the imbalanced control strengths of subject and motion hinder their performance, leading to trade-offs where enhancing one aspect degrades another.
In contrast, our \frameworkplain harmoniously generates customized videos with desired subject appearances and motion movements under various contexts.
Furthermore, our method effectively constrains subjects within the bounding boxes, better aligning with user preferences and improving real-world applicability.

The quantitative comparison results are presented in Tab.~\ref{tab:combine_compare}.
Our \frameworkplain consistently surpasses all baseline methods in text alignment, subject fidelity, and motion control precision, while achieving comparable Temporal Consistency.
Notably, our approach significantly outperforms the baselines in the mIoU and CD metrics, verifying our robust motion control capabilities. 
In contrast, DreamVideo shows the second-best CLIP-I and DINO-I scores but inferior mIoU and CD, indicating its strength in preserving subject identity despite limitations in motion movements.
MotionBooth exhibits the lowest CLIP-T due to the fine-tuning of the whole model,
but achieves better mIoU and CD metrics than DreamVideo, suggesting that using explicit motion control signals (\eg, bounding boxes) may be more effective than learning from the reference video.

\begin{table}[t]
    \centering
    \resizebox{\columnwidth}{!}{
    \begin{tabular}{cccccccccc} 
        \textbf{Method} & 
        \textbf{CLIP-T}& \textbf{R-CLIP}& \textbf{R-DINO} & \textbf{CLIP-I}& \textbf{DINO-I} & \tabincell{c}{\textbf{T. Cons.} } & \textbf{mIoU} & \textbf{CD} $\downarrow$ \\ 
        \shline
        \noalign{\smallskip}
         DreamVideo & 0.289 & 0.682 & 0.244 & 0.692 & 0.386 & 0.966 & 0.169 & 0.196 \\
         MotionBooth & 0.267 & 0.708 & 0.301 & 0.686 & 0.383 & \textbf{0.970} & 0.351 & 0.097 \\
         \hline
         \noalign{\smallskip}
         \textbf{\framework} & \textbf{0.303} & \textbf{0.751} & \textbf{0.392} & \textbf{0.694} & \textbf{0.411} & 0.968 & \textbf{0.670} & \textbf{0.048} \\ 
    \end{tabular}
    }
    \caption{\textbf{Quantitative comparison of joint subject customization and motion control.}
    }
    \label{tab:combine_compare}
\end{table}
\begin{figure}[t]
  \centering
  \includegraphics[width=1.0\linewidth]{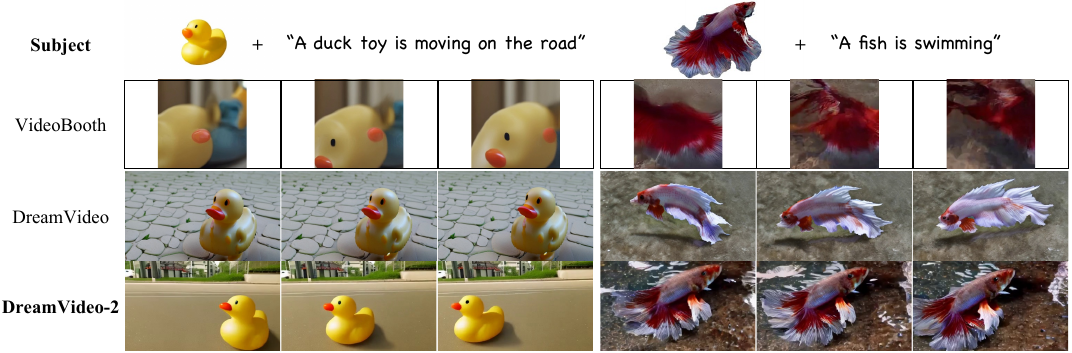}
  \caption{\textbf{Qualitative comparison of subject customization}. \frameworkplain generates videos with accurate subject appearance and enhanced motion dynamics, aligning with provided prompts.}
  \label{fig:compare_id}
\end{figure}

\begin{wraptable}{r}{0.5\textwidth}
    \centering
    \vspace{-4mm}
    \resizebox{\linewidth}{!}{
    \begin{tabular}{ccccccc} 
        \textbf{Method} & 
        \textbf{CLIP-T}& \textbf{CLIP-I}& \textbf{DINO-I} & \tabincell{c}{\textbf{T. Cons.} } & \textbf{DD} \\ 
        \shline
        \noalign{\smallskip}
        DreamVideo & 0.290 & 0.714 & 0.470 & \textbf{0.975} & 0.592 \\
         VideoBooth & 0.274 & \textbf{0.724} & 0.459 & 0.970 & 0.780 \\
         \hline
         \noalign{\smallskip}
         \textbf{\framework} & \textbf{0.297} & 0.721 & \textbf{0.472} & 0.972 & \textbf{0.952} \\ 
    \end{tabular}
    }
    \caption{\textbf{Quantitative comparison of subject customization.}}
    \vspace{-1em}
    \label{tab:id_compare}
\end{wraptable}
\textbf{Subject customization.}\quad
We evaluate the independent subject customization capabilities.
Fig.~\ref{fig:compare_id} presents qualitative comparison results.
We observe that VideoBooth exhibits limited generalization for subjects not included in its training data,
while DreamVideo fails to capture appearance details when trained on a single image.
In contrast, when trained on the same dataset as VideoBooth, our \frameworkplain with reference attention and reweighted diffusion loss generates videos with desired subjects while conforming to textual prompts.

Tab.~\ref{tab:id_compare} shows the quantitative comparison results. 
While \frameworkplain remains comparable CLIP-I and Temporal Consistency, it achieves the highest CLIP-T, DINO-I, and Dynamic Degree, verifying the superior of our method in text alignment, subject fidelity, and motion dynamics.
\begin{figure}[t]
  \centering
  \includegraphics[width=1.0\linewidth]{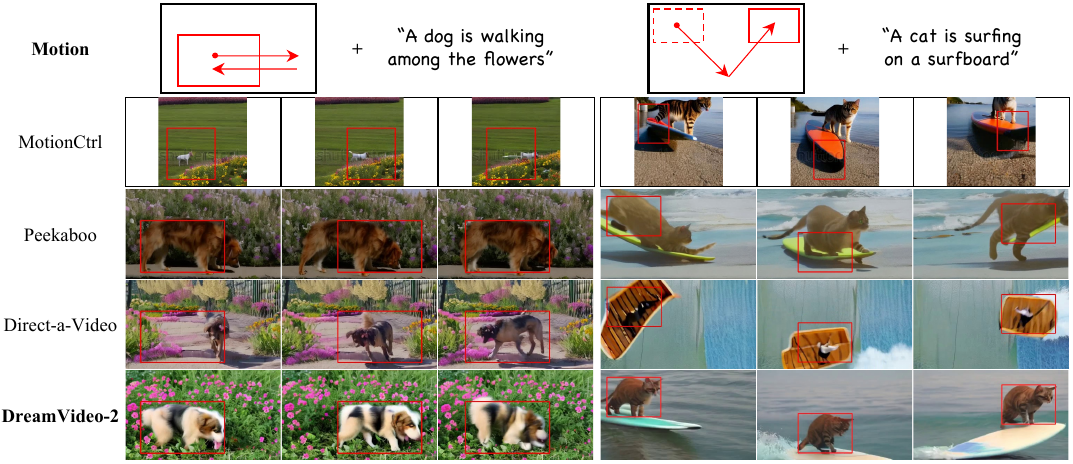}
  \caption{\textbf{Qualitative comparison of motion control}.
  Our \frameworkplain achieves precise motion trajectory control and effectively maintains subjects within the specified bounding boxes.}
  \label{fig:compare_motion}
\end{figure}

\begin{wraptable}{r}{0.5\textwidth}
    \centering
    \resizebox{\linewidth}{!}{
    \begin{tabular}{cccccc} 
        \textbf{Method} &  
        \textbf{CLIP-T}& \tabincell{c}{\textbf{T. Cons.} } & \textbf{mIoU} & \textbf{CD} $\downarrow$ \\ 
        \shline
        \noalign{\smallskip}
         Peekaboo & 0.318 & 0.968 & 0.322 & 0.117  \\
         Direct-a-Video & 0.312 & 0.965 & 0.355 & 0.124 \\
         MotionCtrl & 0.321 & \textbf{0.971} & 0.248 & 0.122 \\
         \hline
         \noalign{\smallskip}
         \textbf{\framework} & \textbf{0.322} & 0.969 & \textbf{0.752} & \textbf{0.039} \\ 
    \end{tabular}
    }
    \caption{\textbf{Quantitative comparison of motion control.}}
    \vspace{-1em}
    \label{tab:motion_compare}
\end{wraptable}
\textbf{Motion control.}\quad
Besides subject customization, we also evaluate the motion control capabilities, as
shown in Fig.~\ref{fig:compare_motion}.
The results suggest that all baselines struggle to accurately control subject movements as defined by bounding boxes.
Meanwhile, Direct-a-Video may generate videos with corrupted object appearances due to its manipulation of attention map values.
In contrast, \frameworkplain with only motion encoder achieves precise motion control and effectively ensures subjects remain within the bounding boxes, demonstrating robust control capabilities.

As shown in Tab.~\ref{tab:motion_compare}, our method, while exhibiting a slightly lower T. Cons. compared to MotionCtrl, achieves the highest CLIP-T and substantially outperforms baselines in both mIoU and CD metrics.

\textbf{User study.}\quad
We conduct user studies to further evaluate our \frameworkplain. 
We ask 15 annotators to rate 300 groups of videos generated by three methods. Each group contains 3 generated videos, a subject image, a textual prompt, and corresponding bounding boxes.
We evaluate all methods with a majority vote from four aspects: Text Alignment, Subject Fidelity, Motion Alignment, and Overall Quality.
Results in Fig.~\ref{fig:user_study} indicate that our method is most preferred by users across four aspects; see Appendix~\ref{app:user_study} for more details of user study.

\subsection{Ablation Studies}
\begin{wrapfigure}{r}{0.5\textwidth}
\centering
\includegraphics[width=1\linewidth]{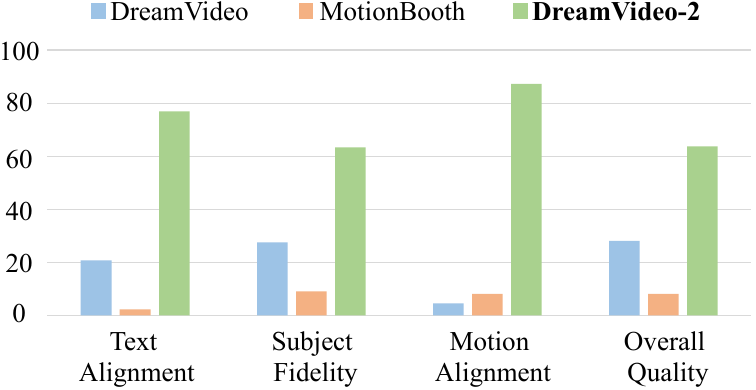}
\caption{\textbf{Human evaluation} on joint subject customization and motion control. 
}
\label{fig:user_study}
\vspace{-1em}
\end{wrapfigure}
\begin{figure}[t]
  \centering
  \includegraphics[width=1.0\linewidth]{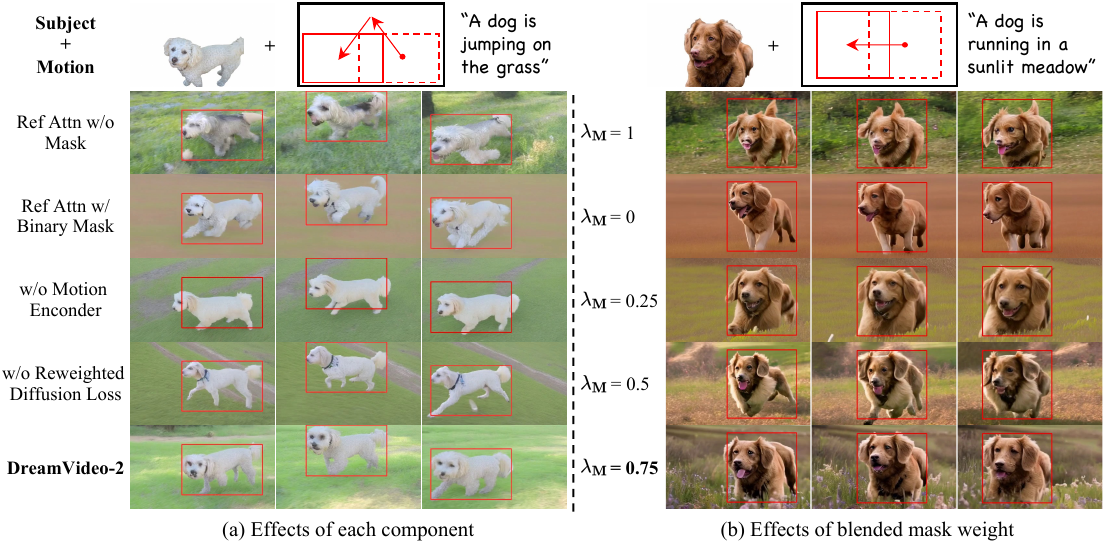}
  \caption{\textbf{Qualitative ablation studies} on each component and blended mask weight.}
  \label{fig:ablation_component_mask}
\end{figure}

\textbf{Effects of each component.}\quad
We perform an ablation study on the effects of each component, as shown in Fig.~\ref{fig:ablation_component_mask}(a). 
We observe that without the mask mechanism or the reweighted diffusion loss, the quality of subject identity degrades due to the dominance of motion control.
While employing binary masks in masked reference attention helps retain subject identity, it often results in a blurry background and low-quality video due to ignoring the background information in attention.
Notably, without the motion encoder, our masked reference attention still achieves rough trajectory control.

Quantitative results in Tab.~\ref{tab:ablation_component} demonstrate that removing the mask mechanism, motion encoder, or reweighted diffusion loss consistently degrades performance across all metrics.
This confirms that each component contributes to the overall performance; see Appendix~\ref{app:ablation} for more ablation studies.

\begin{table}[t]
    \centering
    \resizebox{\columnwidth}{!}{
    \begin{tabular}{lccccccccc} 
         &
        \textbf{CLIP-T}& \textbf{R-CLIP}& \textbf{R-DINO} & \textbf{CLIP-I}& \textbf{DINO-I} & \tabincell{c}{\textbf{T. Cons.} } & \textbf{mIoU} & \textbf{CD} $\downarrow$ \\ 
        \shline
        \noalign{\smallskip}
         Ref Attn w/o Mask ($\lambda_{\mathbf{M}}$ = 1)  & 0.301 & 0.744 & 0.370 & 0.682 & 0.375 & 0.963 & 0.601 & 0.055 \\
         Ref Attn w/ Binary Mask ($\lambda_{\mathbf{M}}$ = 0) & 0.293 & \textbf{0.755} & \underline{0.388} & \textbf{0.696} & 0.394 & \underline{0.967} & \textbf{0.706} & \underline{0.044} \\
         Ref Attn w/ Blended Mask ($\lambda_{\mathbf{M}}$ = 0.25) & 0.299 & 0.748 & 0.379 & 0.685 & \underline{0.395} & 0.964 & \underline{0.693} & \textbf{0.041} \\
         Ref Attn w/ Blended Mask ($\lambda_{\mathbf{M}}$ = 0.5) & 0.301 & 0.748 & 0.376 & \underline{0.694} & 0.386 & 0.961 & 0.664 & 0.051 \\
         w/o Motion Encoder & \underline{0.302} & 0.731 & 0.325 & 0.690 & 0.389 & 0.963 & 0.587 & 0.062 \\
         w/o Reweighted Diffusion Loss & 0.300 & 0.740 & 0.362 & 0.673 & 0.382 & 0.961 & 0.650 & 0.053 \\
         \hline
         \noalign{\smallskip}
         \textbf{\framework} ($\lambda_{\mathbf{M}}$ = 0.75) & \textbf{0.303} & \underline{0.751} & \textbf{0.392} & \underline{0.694} & \textbf{0.411} & \textbf{0.968} & 0.670 & 0.048 \\ 
    \end{tabular}
    }
    \caption{\textbf{Quantitative ablation studies} on each component and blended mask weight.
    }
    \label{tab:ablation_component}
\end{table}

\textbf{Effects of blended mask weight $\lambda_{\mathbf{M}}$.}\quad
To determine the optimal blended mask weight $\lambda_{\mathbf{M}}$, we vary its value and measure its impact.
As shown in Fig.~\ref{fig:ablation_component_mask}(b), using $\lambda_{\mathbf{M}}=1$ results in a degradation of subject identity, while $\lambda_{\mathbf{M}}=0$ leads to blurred backgrounds.
We also observe that increasing $\lambda_{\mathbf{M}}$ can enhance video quality.
To balance subject identity and video quality, we finalize on $\lambda_{\mathbf{M}}=0.75$.

Tab.~\ref{tab:ablation_component} shows the quantitative results. 
$\lambda_{\mathbf{M}}=0$ causes the worst CLIP-T but the highest mIoU.
We argue that a smaller $\lambda_{\mathbf{M}}$ enhances positional information but suppresses background, resulting in improved control precision but degraded video quality.
Additionally, results indicate that using blended masks consistently outperforms its absence in subject fidelity, underscoring its efficacy.
\section{Conclusion}
\label{sec:conclusion}
In this paper, we present \frameworkplain, a novel zero-shot video customization framework that generates videos with specified subjects and motion trajectories.
We introduce reference attention for subject learning and devise a mask-guided motion module 
for motion control.
To address the problem of motion control dominance in \frameworkplain, we 
introduce blended masks into reference attention and design a reweighted diffusion loss, effectively balancing subject learning and motion control.
Extensive experimental results on our newly curated video dataset demonstrate the superiority of \frameworkplain in both subject customization and motion trajectory control.
\\
\textbf{Limitations.}\quad
Although our method can customize a single subject with a single trajectory, it fails to generate videos containing multiple subjects and trajectories. 
One solution is to construct a more diverse dataset and train a general model.
We provide more discussions in Appendix~\ref{app:limit}.
\section{Ethics Statement}
Unlike previous video customization methods that require complicated test-time fine-tuning, our approach enables users to flexibly create customized videos featuring specified subjects and motion trajectories, without the need for fine-tuning or manipulation during inference.
This tuning-free paradigm significantly enhances the real-world applications of customized video generation.
Nonetheless, our method still encounters challenges common to generative models, such as the potential for creating fake data. Implementing robust video forgery detection techniques may address these concerns.
In addition, we commit to adhering to ethical guidelines when releasing our dataset.

\section{Reproducibility Statement}
We make the following efforts to ensure the reproducibility of \frameworkplain: 
(1) Our dataset, code, and trained model weights will be made publicly available. 
(2) We provide the complete descriptions of the dataset construction pipeline in Appendix~\ref{app:dataset_construction}.
(3) We provide implementation details in Sec.~\ref{sec:exp_setup} and Appendix~\ref{app:exp_details}. 
(4) We present the details of the human evaluation setups in Appendix~\ref{app:user_study}.

\clearpage
\appendix
\section{Appendix}
\subsection{Dataset Construction}
\label{app:dataset_construction}
To facilitate the task of zero-shot video customization with subject and motion control,
we curate a single-subject video dataset that encompasses video captions, video masks, and bounding boxes from the WebVid-10M~\citep{webvid10m} dataset and our internal data.
The WebVid-10M dataset comprises 10 million video-text data pairs and is widely used for text-to-video generation. 

\begin{figure}[h]
  \centering
  \includegraphics[width=1.0\linewidth]{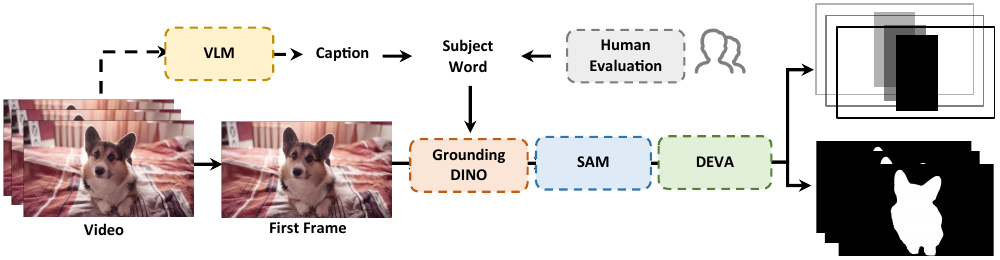}
  \caption{\textbf{Pipeline of dataset construction}.
  }
  \label{fig:dataset_construction}
\end{figure}

We obtain comprehensive annotations by segmenting the subjects of all frames for each video using the Grounding DINO~\citep{groundingdino}, SAM~\citep{sam}, and DEVA~\citep{cheng2023tracking} models, as shown in Fig.~\ref{fig:dataset_construction}.
Specifically, we first extract noun chunks as the initial subject word from the video caption using the spaCy and NLTK libraries. 
For videos that lack the caption, we use a pre-trained Visual Language Model~\citep{lin2024vila} to get its textual description.
Then, we use the NLTK library to perform lemmatization and filter out non-words while asking some annotators to refine the subject words.
Subsequently, we generate the first frame's bounding boxes using Grounding DINO based on the subject word and feed the bounding boxes into SAM to get the subject mask.
We then utilize the object tracker DEVA to populate the mask across all frames of the video, thereby acquiring bounding boxes and masks for all frames.
 
Since we focus on single-subject video customization, we filter out videos that contain multiple subjects for the subject word by the number of bounding boxes in the first frame.
We also filter out subjects that are either too small or too large (\textit{i.e.}, those nearly matching the size of the entire video) by assessing the ratio of the width, height, and area of the subject’s bounding box to the entire video.
Furthermore, we observe a considerable proportion of WebVid-10M videos lacking substantial subject movements.
To ensure the motion dynamic of our dataset, we evaluate each video in the WebVid-10M dataset by comparing their bounding boxes of the first and last frames, retaining those clips where sufficient differences exist between these frames.

After data filtering, we obtain 230,160 video data pairs and 2,538 object classes in the current version. 
The detailed comparison of our dataset with related video datasets is summarized in Tab.~\ref{tab:app_dataset_compare}.
We will further process the WebVid-10M dataset and incorporate more filtered data into our dataset.

\subsection{Experimental Details}
\label{app:exp_details}
\textbf{Baselines.}\quad
Since ModelScopeT2V~\citep{modelScope} generates videos at a resolution of 256$\times$256 and exhibits relatively low quality, we adopt ZeroScope, which is further trained on ModelScopeT2V with additional data to produce videos at a resolution of 576$\times$320, as the base model for all baselines except VideoBooth~\citep{jiang2024videobooth} and MotionCtrl~\citep{wang2024motionctrl}, which utilize their collected datasets to train their own models. 
We follow the default hyperparameter settings from baseline papers for all comparison experiments.

For the task of simultaneously controlling subject appearances and motions, there are currently two methods for us to compare: DreamVideo~\citep{wei2024dreamvideo} and MotionBooth~\citep{wu2024motionbooth}, both requiring fine-tuning at inference time.
Since DreamVideo takes reference videos instead of bounding boxes as motion control signals, we use the video corresponding to the bounding boxes from the DAVIS~\citep{davis} dataset for training DreamVideo's motion adapter.

In addition, we evaluate the performance of independent subject customization or motion control.
For subject customization, we compare our method to DreamVideo and VideoBooth.
Since VideoBooth is also a tuning-free framework, we train our \frameworkplain without the motion encoder and blended mask mechanism, using the same dataset as VideoBooth for a fair comparison.
For motion control, we compare our approach with Peekaboo~\citep{jain2024peekaboo}, Direct-a-Video~\cite{direct_a_video} and MotionCtrl~\citep{wang2024motionctrl}.
Both Peekaboo and Direct-a-Video are training-free methods, while MotionCtrl curates a dataset containing 243,000 videos to train its object motion control module.
Here, we only train the motion encoder in \frameworkplain to enable motion control.

\textbf{Evaluation metrics.}\quad
We detail the use of 9 metrics mentioned in the main paper as follows:
\textbf{1)} 
For overall consistency, we employ
CLIP image-text similarity (CLIP-T), Temporal Consistency (T. Cons.)~\citep{gen1}, and Dynamic Degree (DD)~\citep{huang2024vbench} metrics.
CLIP-T calculates the average cosine similarity between CLIP~\citep{clip} image embeddings of all generated frames and their text embedding.
T. Cons. computes the average cosine similarity across all pairs of consecutive generated frames.
DD uses optical flow to measure the motion intensity, following VBench~\citep{huang2024vbench}.
\textbf{2)}
For subject fidelity, we introduce four metrics: CLIP image similarity (CLIP-I), DINO image similarity (DINO-I), region CLIP-I (R-CLIP), and region DINO-I (R-DINO) metrics~\citep{dreambooth, wei2024dreamvideo, wu2024motionbooth}.
CLIP-I and DINO-I use the CLIP model and 
ViTS/16 
DINO~\cite{dino} model to compute the 
average cosine 
similarities between the subject image and generated frames, respectively.
Furthermore, since we focus on subjects appearing in desired positions, we adopt 
R-CLIP and R-DINO metrics to evaluate the region subject fidelity, following~\citep{wu2024motionbooth}.
R-CLIP and R-DINO compute the similarities between the subject image and frame regions defined by bounding boxes.
\textbf{3)}
For motion control precision, we use the Mean Intersection of Union (mIoU) and Centroid Distance (CD) metrics~\citep{qiu2024freetraj}.
mIoU calculates the average overlap between predicted and ground truth bounding boxes.
CD computes the normalized distance between the centroid of the generated subject and target bounding boxes.

\subsection{More Ablation Studies}
\label{app:ablation}
\textbf{Effects of reweighted diffusion loss weight $\lambda_{\mathcal{L}}$.}\quad
To evaluate the effects of reweighted diffusion loss weight on performance, we test various values of $\lambda_{\mathcal{L}}$, as summarized in Tab.~\ref{tab:ablation_loss_weight}.
Our results indicate that without using reweighted diffusion loss (\ie, $\lambda_{\mathcal{L}}$=1) results in the poorest performance across most metrics. Increasing $\lambda_{\mathcal{L}}$ to 1.5 or 2 yields improvements in all metrics, confirming that enhancing the loss weight of regions inside bounding boxes during training strengthens subject identity. 
On the other hand, setting $\lambda_{\mathcal{L}}$ too high (\eg, $\lambda_{\mathcal{L}}$ = 4) does not improve subject fidelity metrics but negatively affects motion control metrics such as mIoU and CD. 
Therefore, we select $\lambda_{\mathcal{L}}$ = 2 for our training.

\begin{table}[h]
    \centering
    \begin{tabular}{cccccccccc} 
        \textbf{$\lambda_{\mathcal{L}}$} & 
        \textbf{CLIP-T}& \textbf{R-CLIP}& \textbf{R-DINO} & \textbf{CLIP-I}& \textbf{DINO-I} & \tabincell{c}{\textbf{T. Cons.} } & \textbf{mIoU} & \textbf{CD} $\downarrow$ \\ 
        \shline
        \noalign{\smallskip}
         1 & 0.300 & 0.740 & 0.362 & 0.673 & 0.382 & 0.961 & 0.650 & 0.053 \\
         1.5 & \underline{0.302} & 0.745 & 0.370 & 0.687 & 0.385 & \underline{0.965} & \textbf{0.676} & \underline{0.050} \\
         2 & \textbf{0.303} & \textbf{0.751} & \textbf{0.392} & \textbf{0.694} & \textbf{0.411} & \textbf{0.968} & \underline{0.670} & \textbf{0.048} \\
         4 & 0.298 & \underline{0.750} & \underline{0.389} & \underline{0.693} & \underline{0.399} & 0.964 & 0.647 & 0.056 \\
    \end{tabular}
    \caption{\textbf{Ablation study on reweighted diffusion loss weight $\lambda_{\mathcal{L}}$.}}
    \label{tab:ablation_loss_weight}
\end{table}

\subsection{More Results}
\textbf{Details about the user study.}\quad
\label{app:user_study}
We conduct a user study involving 20 subjects and 15 motion trajectories, generating 300 videos per method using randomly selected textual prompts.
Participants are presented with four sets of questions for each of the three anonymous methods, paired with one reference image and one bounding box sequence indicating motion trajectory.
Given the three generated videos in each group, we ask each participant the following questions: 
(1) Text Alignment:
``Which video better matches the text description?'';
(2) Subject Fidelity: 
``Which video’s subject is more similar to the target subject?''; 
(3) Motion Alignment: 
``Which video’s subject movement is more consistent with the target trajectory?'';
and (4) Overall Quality:
``Which video exhibits better quality and minimal flicker?''.
Results of the user study are illustrated in Fig.~\ref{fig:user_study}.

\begin{figure}[t]
  \centering
  \includegraphics[width=1.0\linewidth]{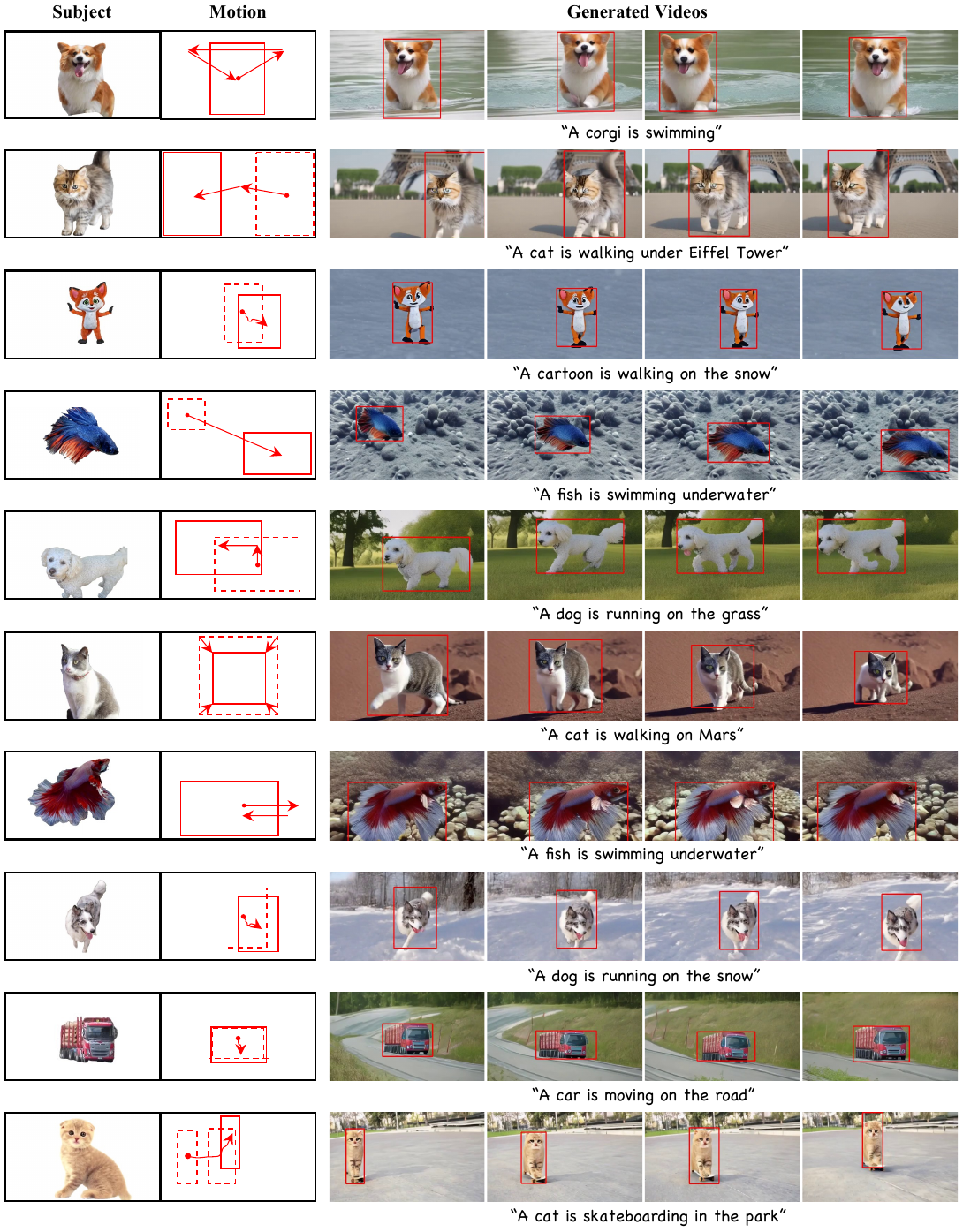}
  \caption{\textbf{More qualitative results of \frameworkplain}. Zoom in for a better view.}
  \label{fig:more_results}
\end{figure}

\textbf{More qualitative results.}\quad
We showcase more results of joint subject customization and motion control in Fig.~\ref{fig:more_results}, providing further evidence of the superiority of our \frameworkplain.

\subsection{Limitations and Future Works}
\label{app:limit}
\begin{wrapfigure}{r}{0.5\textwidth}
\centering
\vspace{-4mm}
\includegraphics[width=1\linewidth]{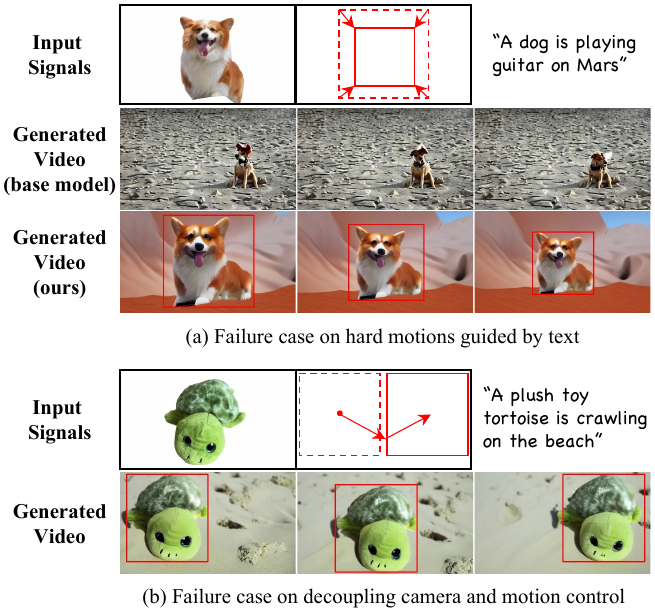}
\caption{\textbf{Failure cases.} (a) Our method is limited by the base model's inherent capabilities.
(b) Our method struggles to decouple the camera and object motion control.
}
\label{fig:fail_cases}
\end{wrapfigure}
In addition to the limitations mentioned in Sec.~\ref{sec:conclusion}, we also provide several failure cases in Fig.~\ref{fig:fail_cases}.
Since we freeze the original 3D UNet parameters during training, our approach is limited by the base model's inherent capabilities, and may fail to generate some rare motions that the subject is unlikely to exhibit.
For example, in Fig.~\ref{fig:fail_cases}(a), the basic model fails to generate a video like ``a dog is playing guitar on Mars'', causing our method to inherit this limitation. 
Employing more advanced T2V models could mitigate this issue.
Another limitation is that our method struggles with decoupling camera and object motion control.
As shown in Fig.~\ref{fig:fail_cases}(b), the model may generate videos with moving cameras and static subjects.
Training a camera control module on dedicated camera movement datasets could aid in addressing this challenge~\citep{wang2024motionctrl, direct_a_video, imageConductor}.

Future work will focus on overcoming these limitations by leveraging a more powerful base T2V model and separating camera movement from our training dataset, aiming for improved performance in real-world applications.

\begin{thebibliography}{103}
\providecommand{\natexlab}[1]{#1}
\providecommand{\url}[1]{\texttt{#1}}
\expandafter\ifx\csname urlstyle\endcsname\relax
  \providecommand{\doi}[1]{doi: #1}\else
  \providecommand{\doi}{doi: \begingroup \urlstyle{rm}\Url}\fi

\bibitem[An et~al.(2023)An, Zhang, Yang, Gupta, Huang, Luo, and
  Yin]{latent_shift}
Jie An, Songyang Zhang, Harry Yang, Sonal Gupta, Jia-Bin Huang, Jiebo Luo, and
  Xi~Yin.
\newblock Latent-shift: Latent diffusion with temporal shift for efficient
  text-to-video generation.
\newblock \emph{arXiv preprint arXiv:2304.08477}, 2023.

\bibitem[Bain et~al.(2021)Bain, Nagrani, Varol, and Zisserman]{webvid10m}
Max Bain, Arsha Nagrani, G{\"u}l Varol, and Andrew Zisserman.
\newblock Frozen in time: A joint video and image encoder for end-to-end
  retrieval.
\newblock In \emph{Proceedings of the IEEE/CVF international conference on
  computer vision}, pp.\  1728--1738, 2021.

\bibitem[Bar-Tal et~al.(2024)Bar-Tal, Chefer, Tov, Herrmann, Paiss, Zada,
  Ephrat, Hur, Li, Michaeli, et~al.]{bar2024lumiere}
Omer Bar-Tal, Hila Chefer, Omer Tov, Charles Herrmann, Roni Paiss, Shiran Zada,
  Ariel Ephrat, Junhwa Hur, Yuanzhen Li, Tomer Michaeli, et~al.
\newblock Lumiere: A space-time diffusion model for video generation.
\newblock \emph{arXiv preprint arXiv:2401.12945}, 2024.

\bibitem[Blattmann et~al.(2023{\natexlab{a}})Blattmann, Dockhorn, Kulal,
  Mendelevitch, Kilian, Lorenz, Levi, English, Voleti, Letts, et~al.]{svd}
Andreas Blattmann, Tim Dockhorn, Sumith Kulal, Daniel Mendelevitch, Maciej
  Kilian, Dominik Lorenz, Yam Levi, Zion English, Vikram Voleti, Adam Letts,
  et~al.
\newblock Stable video diffusion: Scaling latent video diffusion models to
  large datasets.
\newblock \emph{arXiv preprint arXiv:2311.15127}, 2023{\natexlab{a}}.

\bibitem[Blattmann et~al.(2023{\natexlab{b}})Blattmann, Rombach, Ling,
  Dockhorn, Kim, Fidler, and Kreis]{align-your-latents}
Andreas Blattmann, Robin Rombach, Huan Ling, Tim Dockhorn, Seung~Wook Kim,
  Sanja Fidler, and Karsten Kreis.
\newblock Align your latents: High-resolution video synthesis with latent
  diffusion models.
\newblock In \emph{Proceedings of the IEEE/CVF Conference on Computer Vision
  and Pattern Recognition}, pp.\  22563--22575, 2023{\natexlab{b}}.

\bibitem[Brooks et~al.(2024)Brooks, Peebles, Holmes, DePue, Guo, Jing, Schnurr,
  Taylor, Luhman, Luhman, Ng, Wang, and Ramesh]{sora}
Tim Brooks, Bill Peebles, Connor Holmes, Will DePue, Yufei Guo, Li~Jing, David
  Schnurr, Joe Taylor, Troy Luhman, Eric Luhman, Clarence Ng, Ricky Wang, and
  Aditya Ramesh.
\newblock Video generation models as world simulators.
\newblock 2024.
\newblock URL
  \url{https://openai.com/research/video-generation-models-as-world-simulators}.

\bibitem[Caron et~al.(2021)Caron, Touvron, Misra, J{\'e}gou, Mairal,
  Bojanowski, and Joulin]{dino}
Mathilde Caron, Hugo Touvron, Ishan Misra, Herv{\'e} J{\'e}gou, Julien Mairal,
  Piotr Bojanowski, and Armand Joulin.
\newblock Emerging properties in self-supervised vision transformers.
\newblock In \emph{Proceedings of the IEEE/CVF international conference on
  computer vision}, pp.\  9650--9660, 2021.

\bibitem[Chefer et~al.(2024)Chefer, Zada, Paiss, Ephrat, Tov, Rubinstein, Wolf,
  Dekel, Michaeli, and Mosseri]{chefer2024still_moving}
Hila Chefer, Shiran Zada, Roni Paiss, Ariel Ephrat, Omer Tov, Michael
  Rubinstein, Lior Wolf, Tali Dekel, Tomer Michaeli, and Inbar Mosseri.
\newblock Still-moving: Customized video generation without customized video
  data.
\newblock \emph{arXiv preprint arXiv:2407.08674}, 2024.

\bibitem[Chen et~al.(2024{\natexlab{a}})Chen, Shu, Chen, He, Wang, and
  Li]{motion_zero}
Changgu Chen, Junwei Shu, Lianggangxu Chen, Gaoqi He, Changbo Wang, and Yang
  Li.
\newblock Motion-zero: Zero-shot moving object control framework for
  diffusion-based video generation.
\newblock \emph{arXiv preprint arXiv:2401.10150}, 2024{\natexlab{a}}.

\bibitem[Chen et~al.(2023{\natexlab{a}})Chen, Xia, He, Zhang, Cun, Yang, Xing,
  Liu, Chen, Wang, et~al.]{videocrafter1}
Haoxin Chen, Menghan Xia, Yingqing He, Yong Zhang, Xiaodong Cun, Shaoshu Yang,
  Jinbo Xing, Yaofang Liu, Qifeng Chen, Xintao Wang, et~al.
\newblock Videocrafter1: Open diffusion models for high-quality video
  generation.
\newblock \emph{arXiv preprint arXiv:2310.19512}, 2023{\natexlab{a}}.

\bibitem[Chen et~al.(2024{\natexlab{b}})Chen, Zhang, Cun, Xia, Wang, Weng, and
  Shan]{chen2024videocrafter2}
Haoxin Chen, Yong Zhang, Xiaodong Cun, Menghan Xia, Xintao Wang, Chao Weng, and
  Ying Shan.
\newblock Videocrafter2: Overcoming data limitations for high-quality video
  diffusion models.
\newblock In \emph{Proceedings of the IEEE/CVF Conference on Computer Vision
  and Pattern Recognition}, pp.\  7310--7320, 2024{\natexlab{b}}.

\bibitem[Chen et~al.(2023{\natexlab{b}})Chen, Wang, Zeng, Zhang, Zhou, Han, and
  Zhu]{chen2023videodreamer}
Hong Chen, Xin Wang, Guanning Zeng, Yipeng Zhang, Yuwei Zhou, Feilin Han, and
  Wenwu Zhu.
\newblock Videodreamer: Customized multi-subject text-to-video generation with
  disen-mix finetuning.
\newblock \emph{arXiv preprint arXiv:2311.00990}, 2023{\natexlab{b}}.

\bibitem[Chen et~al.(2023{\natexlab{c}})Chen, Zhang, Wang, Duan, Zhou, and
  Zhu]{disenbooth}
Hong Chen, Yipeng Zhang, Xin Wang, Xuguang Duan, Yuwei Zhou, and Wenwu Zhu.
\newblock Disenbooth: Disentangled parameter-efficient tuning for
  subject-driven text-to-image generation.
\newblock \emph{arXiv preprint arXiv:2305.03374}, 2023{\natexlab{c}}.

\bibitem[Chen et~al.(2024{\natexlab{c}})Chen, Wang, Zhang, Zhou, Zhang, Tang,
  and Zhu]{chen2024disenstudio}
Hong Chen, Xin Wang, Yipeng Zhang, Yuwei Zhou, Zeyang Zhang, Siao Tang, and
  Wenwu Zhu.
\newblock Disenstudio: Customized multi-subject text-to-video generation with
  disentangled spatial control.
\newblock \emph{arXiv preprint arXiv:2405.12796}, 2024{\natexlab{c}}.

\bibitem[Chen et~al.(2024{\natexlab{d}})Chen, Hu, Li, Ruiz, Jia, Chang, and
  Cohen]{SuTI}
Wenhu Chen, Hexiang Hu, Yandong Li, Nataniel Ruiz, Xuhui Jia, Ming-Wei Chang,
  and William~W Cohen.
\newblock Subject-driven text-to-image generation via apprenticeship learning.
\newblock \emph{Advances in Neural Information Processing Systems}, 36,
  2024{\natexlab{d}}.

\bibitem[Chen et~al.(2023{\natexlab{d}})Chen, Huang, Liu, Shen, Zhao, and
  Zhao]{anydoor}
Xi~Chen, Lianghua Huang, Yu~Liu, Yujun Shen, Deli Zhao, and Hengshuang Zhao.
\newblock Anydoor: Zero-shot object-level image customization.
\newblock \emph{arXiv preprint arXiv:2307.09481}, 2023{\natexlab{d}}.

\bibitem[Chen et~al.(2024{\natexlab{e}})Chen, Huang, Liu, Shen, Zhao, and
  Zhao]{chen2024anydoor}
Xi~Chen, Lianghua Huang, Yu~Liu, Yujun Shen, Deli Zhao, and Hengshuang Zhao.
\newblock Anydoor: Zero-shot object-level image customization.
\newblock In \emph{Proceedings of the IEEE/CVF Conference on Computer Vision
  and Pattern Recognition}, pp.\  6593--6602, 2024{\natexlab{e}}.

\bibitem[Cheng et~al.(2023)Cheng, Oh, Price, Schwing, and
  Lee]{cheng2023tracking}
Ho~Kei Cheng, Seoung~Wug Oh, Brian Price, Alexander Schwing, and Joon-Young
  Lee.
\newblock Tracking anything with decoupled video segmentation.
\newblock In \emph{Proceedings of the IEEE/CVF International Conference on
  Computer Vision}, pp.\  1316--1326, 2023.

\bibitem[Esser et~al.(2023)Esser, Chiu, Atighehchian, Granskog, and
  Germanidis]{gen1}
Patrick Esser, Johnathan Chiu, Parmida Atighehchian, Jonathan Granskog, and
  Anastasis Germanidis.
\newblock Structure and content-guided video synthesis with diffusion models.
\newblock In \emph{Proceedings of the IEEE/CVF International Conference on
  Computer Vision}, pp.\  7346--7356, 2023.

\bibitem[Gal et~al.(2022)Gal, Alaluf, Atzmon, Patashnik, Bermano, Chechik, and
  Cohen-Or]{textInversion}
Rinon Gal, Yuval Alaluf, Yuval Atzmon, Or~Patashnik, Amit~H Bermano, Gal
  Chechik, and Daniel Cohen-Or.
\newblock An image is worth one word: Personalizing text-to-image generation
  using textual inversion.
\newblock \emph{arXiv preprint arXiv:2208.01618}, 2022.

\bibitem[Gu et~al.(2024)Gu, Wang, Wu, Shi, Chen, Fan, Xiao, Zhao, Chang, Wu,
  et~al.]{mix_of_show}
Yuchao Gu, Xintao Wang, Jay~Zhangjie Wu, Yujun Shi, Yunpeng Chen, Zihan Fan,
  Wuyou Xiao, Rui Zhao, Shuning Chang, Weijia Wu, et~al.
\newblock Mix-of-show: Decentralized low-rank adaptation for multi-concept
  customization of diffusion models.
\newblock \emph{Advances in Neural Information Processing Systems}, 36, 2024.

\bibitem[Guo et~al.(2023{\natexlab{a}})Guo, Yang, Rao, Agrawala, Lin, and
  Dai]{guo2023sparsectrl}
Yuwei Guo, Ceyuan Yang, Anyi Rao, Maneesh Agrawala, Dahua Lin, and Bo~Dai.
\newblock Sparsectrl: Adding sparse controls to text-to-video diffusion models.
\newblock \emph{arXiv preprint arXiv:2311.16933}, 2023{\natexlab{a}}.

\bibitem[Guo et~al.(2023{\natexlab{b}})Guo, Yang, Rao, Wang, Qiao, Lin, and
  Dai]{animatediff}
Yuwei Guo, Ceyuan Yang, Anyi Rao, Yaohui Wang, Yu~Qiao, Dahua Lin, and Bo~Dai.
\newblock Animatediff: Animate your personalized text-to-image diffusion models
  without specific tuning.
\newblock \emph{arXiv preprint arXiv:2307.04725}, 2023{\natexlab{b}}.

\bibitem[Gupta et~al.(2023)Gupta, Yu, Sohn, Gu, Hahn, Fei-Fei, Essa, Jiang, and
  Lezama]{gupta2023photorealistic}
Agrim Gupta, Lijun Yu, Kihyuk Sohn, Xiuye Gu, Meera Hahn, Li~Fei-Fei, Irfan
  Essa, Lu~Jiang, and Jos{\'e} Lezama.
\newblock Photorealistic video generation with diffusion models.
\newblock \emph{arXiv preprint arXiv:2312.06662}, 2023.

\bibitem[Han et~al.(2023)Han, Li, Zhang, Milanfar, Metaxas, and
  Yang]{han2023svdiff}
Ligong Han, Yinxiao Li, Han Zhang, Peyman Milanfar, Dimitris Metaxas, and Feng
  Yang.
\newblock Svdiff: Compact parameter space for diffusion fine-tuning.
\newblock In \emph{Proceedings of the IEEE/CVF International Conference on
  Computer Vision}, pp.\  7323--7334, 2023.

\bibitem[Han et~al.(2024)Han, Zhu, He, Chen, Ge, Li, Li, Zhang, Wang, and
  Liu]{han2024face_adapter}
Yue Han, Junwei Zhu, Keke He, Xu~Chen, Yanhao Ge, Wei Li, Xiangtai Li,
  Jiangning Zhang, Chengjie Wang, and Yong Liu.
\newblock Face adapter for pre-trained diffusion models with fine-grained id
  and attribute control.
\newblock \emph{arXiv preprint arXiv:2405.12970}, 2024.

\bibitem[He et~al.(2024{\natexlab{a}})He, Xu, Guo, Wetzstein, Dai, Li, and
  Yang]{he2024cameractrl}
Hao He, Yinghao Xu, Yuwei Guo, Gordon Wetzstein, Bo~Dai, Hongsheng Li, and
  Ceyuan Yang.
\newblock Cameractrl: Enabling camera control for text-to-video generation.
\newblock \emph{arXiv preprint arXiv:2404.02101}, 2024{\natexlab{a}}.

\bibitem[He et~al.(2024{\natexlab{b}})He, Liu, Qian, Wang, Hu, Cao, Yan, Zhou,
  and Zhang]{he2024id_animator}
Xuanhua He, Quande Liu, Shengju Qian, Xin Wang, Tao Hu, Ke~Cao, Keyu Yan, Man
  Zhou, and Jie Zhang.
\newblock Id-animator: Zero-shot identity-preserving human video generation.
\newblock \emph{arXiv preprint arXiv:2404.15275}, 2024{\natexlab{b}}.

\bibitem[He et~al.(2022)He, Yang, Zhang, Shan, and Chen]{LVDM}
Yingqing He, Tianyu Yang, Yong Zhang, Ying Shan, and Qifeng Chen.
\newblock Latent video diffusion models for high-fidelity long video
  generation.
\newblock \emph{arXiv preprint arXiv:2211.13221}, 2022.

\bibitem[Ho \& Salimans(2022)Ho and Salimans]{ho2022classifier_free_guide}
Jonathan Ho and Tim Salimans.
\newblock Classifier-free diffusion guidance.
\newblock \emph{arXiv preprint arXiv:2207.12598}, 2022.

\bibitem[Ho et~al.(2020)Ho, Jain, and Abbeel]{DDPM}
Jonathan Ho, Ajay Jain, and Pieter Abbeel.
\newblock Denoising diffusion probabilistic models.
\newblock \emph{Advances in neural information processing systems},
  33:\penalty0 6840--6851, 2020.

\bibitem[Ho et~al.(2022{\natexlab{a}})Ho, Chan, Saharia, Whang, Gao, Gritsenko,
  Kingma, Poole, Norouzi, Fleet, et~al.]{imagenVideo}
Jonathan Ho, William Chan, Chitwan Saharia, Jay Whang, Ruiqi Gao, Alexey
  Gritsenko, Diederik~P Kingma, Ben Poole, Mohammad Norouzi, David~J Fleet,
  et~al.
\newblock Imagen video: High definition video generation with diffusion models.
\newblock \emph{arXiv preprint arXiv:2210.02303}, 2022{\natexlab{a}}.

\bibitem[Ho et~al.(2022{\natexlab{b}})Ho, Salimans, Gritsenko, Chan, Norouzi,
  and Fleet]{VDM}
Jonathan Ho, Tim Salimans, Alexey Gritsenko, William Chan, Mohammad Norouzi,
  and David~J. Fleet.
\newblock Video diffusion models.
\newblock \emph{arXiv preprint arXiv:2204.03458}, 2022{\natexlab{b}}.

\bibitem[Hong et~al.(2022)Hong, Ding, Zheng, Liu, and Tang]{hong2022cogvideo}
Wenyi Hong, Ming Ding, Wendi Zheng, Xinghan Liu, and Jie Tang.
\newblock Cogvideo: Large-scale pretraining for text-to-video generation via
  transformers.
\newblock \emph{arXiv preprint arXiv:2205.15868}, 2022.

\bibitem[Hu(2024)]{hu2024animateanyone}
Li~Hu.
\newblock Animate anyone: Consistent and controllable image-to-video synthesis
  for character animation.
\newblock In \emph{Proceedings of the IEEE/CVF Conference on Computer Vision
  and Pattern Recognition}, pp.\  8153--8163, 2024.

\bibitem[Hua et~al.(2023)Hua, Liu, Ding, Liu, Wu, and He]{hua2023dreamtuner}
Miao Hua, Jiawei Liu, Fei Ding, Wei Liu, Jie Wu, and Qian He.
\newblock Dreamtuner: Single image is enough for subject-driven generation.
\newblock \emph{arXiv preprint arXiv:2312.13691}, 2023.

\bibitem[Huang et~al.(2019)Huang, Zhao, and Huang]{GOT10k}
Lianghua Huang, Xin Zhao, and Kaiqi Huang.
\newblock Got-10k: A large high-diversity benchmark for generic object tracking
  in the wild.
\newblock \emph{IEEE transactions on pattern analysis and machine
  intelligence}, 43\penalty0 (5):\penalty0 1562--1577, 2019.

\bibitem[Huang et~al.(2024)Huang, He, Yu, Zhang, Si, Jiang, Zhang, Wu, Jin,
  Chanpaisit, et~al.]{huang2024vbench}
Ziqi Huang, Yinan He, Jiashuo Yu, Fan Zhang, Chenyang Si, Yuming Jiang, Yuanhan
  Zhang, Tianxing Wu, Qingyang Jin, Nattapol Chanpaisit, et~al.
\newblock Vbench: Comprehensive benchmark suite for video generative models.
\newblock In \emph{Proceedings of the IEEE/CVF Conference on Computer Vision
  and Pattern Recognition}, pp.\  21807--21818, 2024.

\bibitem[Jain et~al.(2024)Jain, Nasery, Vineet, and Behl]{jain2024peekaboo}
Yash Jain, Anshul Nasery, Vibhav Vineet, and Harkirat Behl.
\newblock Peekaboo: Interactive video generation via masked-diffusion.
\newblock In \emph{Proceedings of the IEEE/CVF Conference on Computer Vision
  and Pattern Recognition}, pp.\  8079--8088, 2024.

\bibitem[Jeong et~al.(2024)Jeong, Park, and Ye]{jeong2024vmc}
Hyeonho Jeong, Geon~Yeong Park, and Jong~Chul Ye.
\newblock Vmc: Video motion customization using temporal attention adaption for
  text-to-video diffusion models.
\newblock In \emph{Proceedings of the IEEE/CVF Conference on Computer Vision
  and Pattern Recognition}, pp.\  9212--9221, 2024.

\bibitem[Jiang et~al.(2024)Jiang, Wu, Yang, Si, Lin, Qiao, Loy, and
  Liu]{jiang2024videobooth}
Yuming Jiang, Tianxing Wu, Shuai Yang, Chenyang Si, Dahua Lin, Yu~Qiao,
  Chen~Change Loy, and Ziwei Liu.
\newblock Videobooth: Diffusion-based video generation with image prompts.
\newblock In \emph{Proceedings of the IEEE/CVF Conference on Computer Vision
  and Pattern Recognition}, pp.\  6689--6700, 2024.

\bibitem[Kingma \& Welling(2013)Kingma and Welling]{vae}
Diederik~P Kingma and Max Welling.
\newblock Auto-encoding variational bayes.
\newblock \emph{arXiv preprint arXiv:1312.6114}, 2013.

\bibitem[Kirillov et~al.(2023)Kirillov, Mintun, Ravi, Mao, Rolland, Gustafson,
  Xiao, Whitehead, Berg, Lo, et~al.]{sam}
Alexander Kirillov, Eric Mintun, Nikhila Ravi, Hanzi Mao, Chloe Rolland, Laura
  Gustafson, Tete Xiao, Spencer Whitehead, Alexander~C Berg, Wan-Yen Lo, et~al.
\newblock Segment anything.
\newblock In \emph{Proceedings of the IEEE/CVF International Conference on
  Computer Vision}, pp.\  4015--4026, 2023.

\bibitem[Kondratyuk et~al.(2023)Kondratyuk, Yu, Gu, Lezama, Huang, Hornung,
  Adam, Akbari, Alon, Birodkar, et~al.]{kondratyuk2023videopoet}
Dan Kondratyuk, Lijun Yu, Xiuye Gu, Jos{\'e} Lezama, Jonathan Huang, Rachel
  Hornung, Hartwig Adam, Hassan Akbari, Yair Alon, Vighnesh Birodkar, et~al.
\newblock Videopoet: A large language model for zero-shot video generation.
\newblock \emph{arXiv preprint arXiv:2312.14125}, 2023.

\bibitem[Kumari et~al.(2023)Kumari, Zhang, Zhang, Shechtman, and
  Zhu]{customDiffusion}
Nupur Kumari, Bingliang Zhang, Richard Zhang, Eli Shechtman, and Jun-Yan Zhu.
\newblock Multi-concept customization of text-to-image diffusion.
\newblock In \emph{Proceedings of the IEEE/CVF Conference on Computer Vision
  and Pattern Recognition}, pp.\  1931--1941, 2023.

\bibitem[Li et~al.(2023{\natexlab{a}})Li, Liu, Xia, Lin, Wang, Zheng, Yang,
  Zhong, Ren, and He]{li2023few}
Hengjia Li, Yang Liu, Linxuan Xia, Yuqi Lin, Wenxiao Wang, Tu~Zheng, Zheng
  Yang, Xiaohui Zhong, Xiaobo Ren, and Xiaofei He.
\newblock Few-shot hybrid domain adaptation of image generator.
\newblock In \emph{The Twelfth International Conference on Learning
  Representations}, 2023{\natexlab{a}}.

\bibitem[Li et~al.(2024{\natexlab{a}})Li, Liu, Lin, Zhang, Zhao, Zheng, Yang,
  Jiang, Wu, Cai, et~al.]{li2024unihda}
Hengjia Li, Yang Liu, Yuqi Lin, Zhanwei Zhang, Yibo Zhao, Tu~Zheng, Zheng Yang,
  Yuchun Jiang, Boxi Wu, Deng Cai, et~al.
\newblock Unihda: Towards universal hybrid domain adaptation of image
  generators.
\newblock \emph{arXiv preprint arXiv:2401.12596}, 2024{\natexlab{a}}.

\bibitem[Li et~al.(2024{\natexlab{b}})Li, Hou, and Loy]{li2024stylegan}
Xiaoming Li, Xinyu Hou, and Chen~Change Loy.
\newblock When stylegan meets stable diffusion: a w+ adapter for personalized
  image generation.
\newblock In \emph{Proceedings of the IEEE/CVF Conference on Computer Vision
  and Pattern Recognition}, pp.\  2187--2196, 2024{\natexlab{b}}.

\bibitem[Li et~al.(2024{\natexlab{c}})Li, Wang, Zhang, Wang, Yuan, Xie, Zou,
  and Shan]{imageConductor}
Yaowei Li, Xintao Wang, Zhaoyang Zhang, Zhouxia Wang, Ziyang Yuan, Liangbin
  Xie, Yuexian Zou, and Ying Shan.
\newblock Image conductor: Precision control for interactive video synthesis.
\newblock \emph{arXiv preprint arXiv:2406.15339}, 2024{\natexlab{c}}.

\bibitem[Li et~al.(2023{\natexlab{b}})Li, Liu, Wu, Mu, Yang, Gao, Li, and
  Lee]{li2023gligen}
Yuheng Li, Haotian Liu, Qingyang Wu, Fangzhou Mu, Jianwei Yang, Jianfeng Gao,
  Chunyuan Li, and Yong~Jae Lee.
\newblock Gligen: Open-set grounded text-to-image generation.
\newblock In \emph{Proceedings of the IEEE/CVF Conference on Computer Vision
  and Pattern Recognition}, pp.\  22511--22521, 2023{\natexlab{b}}.

\bibitem[Lin et~al.(2024)Lin, Yin, Ping, Molchanov, Shoeybi, and
  Han]{lin2024vila}
Ji~Lin, Hongxu Yin, Wei Ping, Pavlo Molchanov, Mohammad Shoeybi, and Song Han.
\newblock Vila: On pre-training for visual language models.
\newblock In \emph{Proceedings of the IEEE/CVF Conference on Computer Vision
  and Pattern Recognition}, pp.\  26689--26699, 2024.

\bibitem[Liu et~al.(2023{\natexlab{a}})Liu, Zeng, Ren, Li, Zhang, Yang, Li,
  Yang, Su, Zhu, et~al.]{groundingdino}
Shilong Liu, Zhaoyang Zeng, Tianhe Ren, Feng Li, Hao Zhang, Jie Yang, Chunyuan
  Li, Jianwei Yang, Hang Su, Jun Zhu, et~al.
\newblock Grounding dino: Marrying dino with grounded pre-training for open-set
  object detection.
\newblock \emph{arXiv preprint arXiv:2303.05499}, 2023{\natexlab{a}}.

\bibitem[Liu et~al.(2023{\natexlab{b}})Liu, Feng, Zhu, Zhang, Zheng, Liu, Zhao,
  Zhou, and Cao]{cones}
Zhiheng Liu, Ruili Feng, Kai Zhu, Yifei Zhang, Kecheng Zheng, Yu~Liu, Deli
  Zhao, Jingren Zhou, and Yang Cao.
\newblock Cones: Concept neurons in diffusion models for customized generation.
\newblock \emph{arXiv preprint arXiv:2303.05125}, 2023{\natexlab{b}}.

\bibitem[Liu et~al.(2023{\natexlab{c}})Liu, Zhang, Shen, Zheng, Zhu, Feng, Liu,
  Zhao, Zhou, and Cao]{cones2}
Zhiheng Liu, Yifei Zhang, Yujun Shen, Kecheng Zheng, Kai Zhu, Ruili Feng,
  Yu~Liu, Deli Zhao, Jingren Zhou, and Yang Cao.
\newblock Cones 2: Customizable image synthesis with multiple subjects.
\newblock In \emph{Proceedings of the 37th International Conference on Neural
  Information Processing Systems}, pp.\  57500--57519, 2023{\natexlab{c}}.

\bibitem[Loshchilov(2017)]{adamw}
I~Loshchilov.
\newblock Decoupled weight decay regularization.
\newblock \emph{arXiv preprint arXiv:1711.05101}, 2017.

\bibitem[Ma et~al.(2023)Ma, Lewis, and Kleijn]{ma2023trailblazer}
Wan-Duo~Kurt Ma, John~P Lewis, and W~Bastiaan Kleijn.
\newblock Trailblazer: Trajectory control for diffusion-based video generation.
\newblock \emph{arXiv preprint arXiv:2401.00896}, 2023.

\bibitem[Ma et~al.(2024{\natexlab{a}})Ma, Wang, Jia, Chen, Liu, Li, Chen, and
  Qiao]{ma2024latte}
Xin Ma, Yaohui Wang, Gengyun Jia, Xinyuan Chen, Ziwei Liu, Yuan-Fang Li,
  Cunjian Chen, and Yu~Qiao.
\newblock Latte: Latent diffusion transformer for video generation.
\newblock \emph{arXiv preprint arXiv:2401.03048}, 2024{\natexlab{a}}.

\bibitem[Ma et~al.(2024{\natexlab{b}})Ma, Zhou, Yeh, Wang, Li, Yang, Dong,
  Keutzer, and Feng]{ma2024magicme}
Ze~Ma, Daquan Zhou, Chun-Hsiao Yeh, Xue-She Wang, Xiuyu Li, Huanrui Yang, Zhen
  Dong, Kurt Keutzer, and Jiashi Feng.
\newblock Magic-me: Identity-specific video customized diffusion.
\newblock \emph{arXiv preprint arXiv:2402.09368}, 2024{\natexlab{b}}.

\bibitem[Molad et~al.(2023)Molad, Horwitz, Valevski, Acha, Matias, Pritch,
  Leviathan, and Hoshen]{dreamix}
Eyal Molad, Eliahu Horwitz, Dani Valevski, Alex~Rav Acha, Yossi Matias, Yael
  Pritch, Yaniv Leviathan, and Yedid Hoshen.
\newblock Dreamix: Video diffusion models are general video editors.
\newblock \emph{arXiv preprint arXiv:2302.01329}, 2023.

\bibitem[Mou et~al.(2024)Mou, Wang, Xie, Wu, Zhang, Qi, and Shan]{mou2024t2i}
Chong Mou, Xintao Wang, Liangbin Xie, Yanze Wu, Jian Zhang, Zhongang Qi, and
  Ying Shan.
\newblock T2i-adapter: Learning adapters to dig out more controllable ability
  for text-to-image diffusion models.
\newblock In \emph{Proceedings of the AAAI Conference on Artificial
  Intelligence}, volume~38, pp.\  4296--4304, 2024.

\bibitem[Ng et~al.(2022)Ng, Ong, Zheng, Ni, Yeo, and Liu]{Animal_Kingdom}
Xun~Long Ng, Kian~Eng Ong, Qichen Zheng, Yun Ni, Si~Yong Yeo, and Jun Liu.
\newblock Animal kingdom: A large and diverse dataset for animal behavior
  understanding.
\newblock In \emph{Proceedings of the IEEE/CVF Conference on Computer Vision
  and Pattern Recognition (CVPR)}, pp.\  19023--19034, June 2022.

\bibitem[Podell et~al.(2023)Podell, English, Lacey, Blattmann, Dockhorn,
  M{\"u}ller, Penna, and Rombach]{podell2023sdxl}
Dustin Podell, Zion English, Kyle Lacey, Andreas Blattmann, Tim Dockhorn, Jonas
  M{\"u}ller, Joe Penna, and Robin Rombach.
\newblock Sdxl: Improving latent diffusion models for high-resolution image
  synthesis.
\newblock \emph{arXiv preprint arXiv:2307.01952}, 2023.

\bibitem[Pont-Tuset et~al.(2017)Pont-Tuset, Perazzi, Caelles, Arbel{\'a}ez,
  Sorkine-Hornung, and Van~Gool]{davis}
Jordi Pont-Tuset, Federico Perazzi, Sergi Caelles, Pablo Arbel{\'a}ez, Alex
  Sorkine-Hornung, and Luc Van~Gool.
\newblock The 2017 davis challenge on video object segmentation.
\newblock \emph{arXiv preprint arXiv:1704.00675}, 2017.

\bibitem[Qing et~al.(2024)Qing, Zhang, Wang, Wang, Wei, Zhang, Gao, and
  Sang]{higen}
Zhiwu Qing, Shiwei Zhang, Jiayu Wang, Xiang Wang, Yujie Wei, Yingya Zhang,
  Changxin Gao, and Nong Sang.
\newblock Hierarchical spatio-temporal decoupling for text-to-video generation.
\newblock In \emph{Proceedings of the IEEE/CVF Conference on Computer Vision
  and Pattern Recognition}, pp.\  6635--6645, 2024.

\bibitem[Qiu et~al.(2024)Qiu, Chen, Wang, He, Xia, and Liu]{qiu2024freetraj}
Haonan Qiu, Zhaoxi Chen, Zhouxia Wang, Yingqing He, Menghan Xia, and Ziwei Liu.
\newblock Freetraj: Tuning-free trajectory control in video diffusion models.
\newblock \emph{arXiv preprint arXiv:2406.16863}, 2024.

\bibitem[Radford et~al.(2021)Radford, Kim, Hallacy, Ramesh, Goh, Agarwal,
  Sastry, Askell, Mishkin, Clark, et~al.]{clip}
Alec Radford, Jong~Wook Kim, Chris Hallacy, Aditya Ramesh, Gabriel Goh,
  Sandhini Agarwal, Girish Sastry, Amanda Askell, Pamela Mishkin, Jack Clark,
  et~al.
\newblock Learning transferable visual models from natural language
  supervision.
\newblock In \emph{International conference on machine learning}, pp.\
  8748--8763, 2021.

\bibitem[Ren et~al.(2024)Ren, Zhou, Yang, Shi, Liu, Liu, Kwon, and
  Shrivastava]{customize_a_video}
Yixuan Ren, Yang Zhou, Jimei Yang, Jing Shi, Difan Liu, Feng Liu, Mingi Kwon,
  and Abhinav Shrivastava.
\newblock Customize-a-video: One-shot motion customization of text-to-video
  diffusion models.
\newblock \emph{arXiv preprint arXiv:2402.14780}, 2024.

\bibitem[Rombach et~al.(2022)Rombach, Blattmann, Lorenz, Esser, and
  Ommer]{stableDiffusion}
Robin Rombach, Andreas Blattmann, Dominik Lorenz, Patrick Esser, and Bj{\"o}rn
  Ommer.
\newblock High-resolution image synthesis with latent diffusion models.
\newblock In \emph{Proceedings of the IEEE/CVF Conference on Computer Vision
  and Pattern Recognition}, pp.\  10684--10695, 2022.

\bibitem[Ruiz et~al.(2023)Ruiz, Li, Jampani, Pritch, Rubinstein, and
  Aberman]{dreambooth}
Nataniel Ruiz, Yuanzhen Li, Varun Jampani, Yael Pritch, Michael Rubinstein, and
  Kfir Aberman.
\newblock Dreambooth: Fine tuning text-to-image diffusion models for
  subject-driven generation.
\newblock In \emph{Proceedings of the IEEE/CVF Conference on Computer Vision
  and Pattern Recognition}, pp.\  22500--22510, 2023.

\bibitem[Ruiz et~al.(2024)Ruiz, Li, Jampani, Wei, Hou, Pritch, Wadhwa,
  Rubinstein, and Aberman]{ruiz2024hyperdreambooth}
Nataniel Ruiz, Yuanzhen Li, Varun Jampani, Wei Wei, Tingbo Hou, Yael Pritch,
  Neal Wadhwa, Michael Rubinstein, and Kfir Aberman.
\newblock Hyperdreambooth: Hypernetworks for fast personalization of
  text-to-image models.
\newblock In \emph{Proceedings of the IEEE/CVF Conference on Computer Vision
  and Pattern Recognition}, pp.\  6527--6536, 2024.

\bibitem[Shi et~al.(2024)Shi, Xiong, Lin, and Jung]{shi2024instantbooth}
Jing Shi, Wei Xiong, Zhe Lin, and Hyun~Joon Jung.
\newblock Instantbooth: Personalized text-to-image generation without test-time
  finetuning.
\newblock In \emph{Proceedings of the IEEE/CVF Conference on Computer Vision
  and Pattern Recognition}, pp.\  8543--8552, 2024.

\bibitem[Singer et~al.(2022)Singer, Polyak, Hayes, Yin, An, Zhang, Hu, Yang,
  Ashual, Gafni, et~al.]{make-a-video}
Uriel Singer, Adam Polyak, Thomas Hayes, Xi~Yin, Jie An, Songyang Zhang, Qiyuan
  Hu, Harry Yang, Oron Ashual, Oran Gafni, et~al.
\newblock Make-a-video: Text-to-video generation without text-video data.
\newblock \emph{arXiv preprint arXiv:2209.14792}, 2022.

\bibitem[Song et~al.(2020)Song, Meng, and Ermon]{DDIM}
Jiaming Song, Chenlin Meng, and Stefano Ermon.
\newblock Denoising diffusion implicit models.
\newblock \emph{arXiv preprint arXiv:2010.02502}, 2020.

\bibitem[Soomro et~al.(2012)Soomro, Zamir, and Shah]{ucf101}
Khurram Soomro, Amir~Roshan Zamir, and Mubarak Shah.
\newblock Ucf101: A dataset of 101 human actions classes from videos in the
  wild.
\newblock \emph{arXiv preprint arXiv:1212.0402}, 2012.

\bibitem[Tan et~al.(2024)Tan, Gong, Wang, Zhang, Zheng, Zheng, Zheng, Chen, and
  Yang]{tan2024animateX}
Shuai Tan, Biao Gong, Xiang Wang, Shiwei Zhang, Dandan Zheng, Ruobing Zheng,
  Kecheng Zheng, Jingdong Chen, and Ming Yang.
\newblock Animate-x: Universal character image animation with enhanced motion
  representation.
\newblock \emph{arXiv preprint arXiv:2410.10306}, 2024.

\bibitem[Wang et~al.(2024{\natexlab{a}})Wang, Chen, Chen, Huang, Jiang, Wang,
  and Shan]{wang2024fldm}
Chenhui Wang, Tao Chen, Zhihao Chen, Zhizhong Huang, Taoran Jiang, Qi~Wang, and
  Hongming Shan.
\newblock Fldm-vton: Faithful latent diffusion model for virtual try-on.
\newblock \emph{arXiv preprint arXiv:2404.14162}, 2024{\natexlab{a}}.

\bibitem[Wang et~al.(2024{\natexlab{b}})Wang, Zhang, Zou, Zeng, Wei, Yuan, and
  Li]{wang2024boximator}
Jiawei Wang, Yuchen Zhang, Jiaxin Zou, Yan Zeng, Guoqiang Wei, Liping Yuan, and
  Hang Li.
\newblock Boximator: Generating rich and controllable motions for video
  synthesis.
\newblock \emph{arXiv preprint arXiv:2402.01566}, 2024{\natexlab{b}}.

\bibitem[Wang et~al.(2023{\natexlab{a}})Wang, Yuan, Chen, Zhang, Wang, and
  Zhang]{modelScope}
Jiuniu Wang, Hangjie Yuan, Dayou Chen, Yingya Zhang, Xiang Wang, and Shiwei
  Zhang.
\newblock Modelscope text-to-video technical report.
\newblock \emph{arXiv preprint arXiv:2308.06571}, 2023{\natexlab{a}}.

\bibitem[Wang et~al.(2024{\natexlab{c}})Wang, Shen, Liang, Tao, Wan, Zhang, Li,
  and Chen]{motionInversion}
Luozhou Wang, Guibao Shen, Yixun Liang, Xin Tao, Pengfei Wan, Di~Zhang, Yijun
  Li, and Yingcong Chen.
\newblock Motion inversion for video customization.
\newblock \emph{arXiv preprint arXiv:2403.20193}, 2024{\natexlab{c}}.

\bibitem[Wang et~al.(2023{\natexlab{b}})Wang, Yuan, Zhang, Chen, Wang, Zhang,
  Shen, Zhao, and Zhou]{videoComposer}
Xiang Wang, Hangjie Yuan, Shiwei Zhang, Dayou Chen, Jiuniu Wang, Yingya Zhang,
  Yujun Shen, Deli Zhao, and Jingren Zhou.
\newblock Videocomposer: Compositional video synthesis with motion
  controllability.
\newblock \emph{arXiv preprint arXiv:2306.02018}, 2023{\natexlab{b}}.

\bibitem[Wang et~al.(2023{\natexlab{c}})Wang, Zhang, Zhang, Liu, Zhang, Gao,
  and Sang]{wang2023videolcm}
Xiang Wang, Shiwei Zhang, Han Zhang, Yu~Liu, Yingya Zhang, Changxin Gao, and
  Nong Sang.
\newblock Videolcm: Video latent consistency model.
\newblock \emph{arXiv preprint arXiv:2312.09109}, 2023{\natexlab{c}}.

\bibitem[Wang et~al.(2024{\natexlab{d}})Wang, Zhang, Yuan, Qing, Gong, Zhang,
  Shen, Gao, and Sang]{tft2v}
Xiang Wang, Shiwei Zhang, Hangjie Yuan, Zhiwu Qing, Biao Gong, Yingya Zhang,
  Yujun Shen, Changxin Gao, and Nong Sang.
\newblock A recipe for scaling up text-to-video generation with text-free
  videos.
\newblock In \emph{Proceedings of the IEEE/CVF Conference on Computer Vision
  and Pattern Recognition}, pp.\  6572--6582, 2024{\natexlab{d}}.

\bibitem[Wang et~al.(2023{\natexlab{d}})Wang, Chen, Ma, Zhou, Huang, Wang,
  Yang, He, Yu, Yang, et~al.]{wang2023lavie}
Yaohui Wang, Xinyuan Chen, Xin Ma, Shangchen Zhou, Ziqi Huang, Yi~Wang, Ceyuan
  Yang, Yinan He, Jiashuo Yu, Peiqing Yang, et~al.
\newblock Lavie: High-quality video generation with cascaded latent diffusion
  models.
\newblock \emph{arXiv preprint arXiv:2309.15103}, 2023{\natexlab{d}}.

\bibitem[Wang et~al.(2024{\natexlab{e}})Wang, Li, Xie, Zhu, Guo, Dou, and
  Li]{wang2024customvideo}
Zhao Wang, Aoxue Li, Enze Xie, Lingting Zhu, Yong Guo, Qi~Dou, and Zhenguo Li.
\newblock Customvideo: Customizing text-to-video generation with multiple
  subjects.
\newblock \emph{arXiv preprint arXiv:2401.09962}, 2024{\natexlab{e}}.

\bibitem[Wang et~al.(2024{\natexlab{f}})Wang, Yuan, Wang, Li, Chen, Xia, Luo,
  and Shan]{wang2024motionctrl}
Zhouxia Wang, Ziyang Yuan, Xintao Wang, Yaowei Li, Tianshui Chen, Menghan Xia,
  Ping Luo, and Ying Shan.
\newblock Motionctrl: A unified and flexible motion controller for video
  generation.
\newblock In \emph{ACM SIGGRAPH 2024 Conference Papers}, pp.\  1--11,
  2024{\natexlab{f}}.

\bibitem[Wei et~al.(2024)Wei, Zhang, Qing, Yuan, Liu, Liu, Zhang, Zhou, and
  Shan]{wei2024dreamvideo}
Yujie Wei, Shiwei Zhang, Zhiwu Qing, Hangjie Yuan, Zhiheng Liu, Yu~Liu, Yingya
  Zhang, Jingren Zhou, and Hongming Shan.
\newblock Dreamvideo: Composing your dream videos with customized subject and
  motion.
\newblock In \emph{Proceedings of the IEEE/CVF Conference on Computer Vision
  and Pattern Recognition}, pp.\  6537--6549, 2024.

\bibitem[Wei et~al.(2023)Wei, Zhang, Ji, Bai, Zhang, and Zuo]{wei2023elite}
Yuxiang Wei, Yabo Zhang, Zhilong Ji, Jinfeng Bai, Lei Zhang, and Wangmeng Zuo.
\newblock Elite: Encoding visual concepts into textual embeddings for
  customized text-to-image generation.
\newblock In \emph{Proceedings of the IEEE/CVF International Conference on
  Computer Vision}, pp.\  15943--15953, 2023.

\bibitem[Wu et~al.(2024{\natexlab{a}})Wu, Li, Zeng, Zhang, Zhou, Li, Tong, and
  Chen]{wu2024motionbooth}
Jianzong Wu, Xiangtai Li, Yanhong Zeng, Jiangning Zhang, Qianyu Zhou, Yining
  Li, Yunhai Tong, and Kai Chen.
\newblock Motionbooth: Motion-aware customized text-to-video generation.
\newblock \emph{arXiv preprint arXiv:2406.17758}, 2024{\natexlab{a}}.

\bibitem[Wu et~al.(2023)Wu, Chen, Yang, Guo, Li, and Zhang]{lamp}
Ruiqi Wu, Liangyu Chen, Tong Yang, Chunle Guo, Chongyi Li, and Xiangyu Zhang.
\newblock Lamp: Learn a motion pattern for few-shot-based video generation.
\newblock \emph{arXiv preprint arXiv:2310.10769}, 2023.

\bibitem[Wu et~al.(2024{\natexlab{b}})Wu, Zhang, Wang, Zhou, Zheng, Qi, Shan,
  and Li]{wu2024customcrafter}
Tao Wu, Yong Zhang, Xintao Wang, Xianpan Zhou, Guangcong Zheng, Zhongang Qi,
  Ying Shan, and Xi~Li.
\newblock Customcrafter: Customized video generation with preserving motion and
  concept composition abilities.
\newblock \emph{arXiv preprint arXiv:2408.13239}, 2024{\natexlab{b}}.

\bibitem[Xiao et~al.(2023)Xiao, Yin, Freeman, Durand, and Han]{fastcomposer}
Guangxuan Xiao, Tianwei Yin, William~T Freeman, Fr{\'e}do Durand, and Song Han.
\newblock Fastcomposer: Tuning-free multi-subject image generation with
  localized attention.
\newblock \emph{arXiv preprint arXiv:2305.10431}, 2023.

\bibitem[Yang et~al.(2024{\natexlab{a}})Yang, Hou, Huang, Ma, Wan, Zhang, Chen,
  and Liao]{direct_a_video}
Shiyuan Yang, Liang Hou, Haibin Huang, Chongyang Ma, Pengfei Wan, Di~Zhang,
  Xiaodong Chen, and Jing Liao.
\newblock Direct-a-video: Customized video generation with user-directed camera
  movement and object motion.
\newblock In \emph{ACM SIGGRAPH 2024 Conference Papers}, pp.\  1--12,
  2024{\natexlab{a}}.

\bibitem[Yang et~al.(2024{\natexlab{b}})Yang, Teng, Zheng, Ding, Huang, Xu,
  Yang, Hong, Zhang, Feng, et~al.]{yang2024cogvideox}
Zhuoyi Yang, Jiayan Teng, Wendi Zheng, Ming Ding, Shiyu Huang, Jiazheng Xu,
  Yuanming Yang, Wenyi Hong, Xiaohan Zhang, Guanyu Feng, et~al.
\newblock Cogvideox: Text-to-video diffusion models with an expert transformer.
\newblock \emph{arXiv preprint arXiv:2408.06072}, 2024{\natexlab{b}}.

\bibitem[Yatim et~al.(2024)Yatim, Fridman, Bar-Tal, Kasten, and Dekel]{DMT}
Danah Yatim, Rafail Fridman, Omer Bar-Tal, Yoni Kasten, and Tali Dekel.
\newblock Space-time diffusion features for zero-shot text-driven motion
  transfer.
\newblock In \emph{Proceedings of the IEEE/CVF Conference on Computer Vision
  and Pattern Recognition}, pp.\  8466--8476, 2024.

\bibitem[Ye et~al.(2023)Ye, Zhang, Liu, Han, and Yang]{ye2023ip_adapter}
Hu~Ye, Jun Zhang, Sibo Liu, Xiao Han, and Wei Yang.
\newblock Ip-adapter: Text compatible image prompt adapter for text-to-image
  diffusion models.
\newblock \emph{arXiv preprint arXiv:2308.06721}, 2023.

\bibitem[Yin et~al.(2023)Yin, Wu, Liang, Shi, Li, Ming, and
  Duan]{yin2023dragnuwa}
Shengming Yin, Chenfei Wu, Jian Liang, Jie Shi, Houqiang Li, Gong Ming, and Nan
  Duan.
\newblock Dragnuwa: Fine-grained control in video generation by integrating
  text, image, and trajectory.
\newblock \emph{arXiv preprint arXiv:2308.08089}, 2023.

\bibitem[Yuan et~al.(2024)Yuan, Zhang, Wang, Wei, Feng, Pan, Zhang, Liu,
  Albanie, and Ni]{yuan2024instructvideo}
Hangjie Yuan, Shiwei Zhang, Xiang Wang, Yujie Wei, Tao Feng, Yining Pan, Yingya
  Zhang, Ziwei Liu, Samuel Albanie, and Dong Ni.
\newblock Instructvideo: instructing video diffusion models with human
  feedback.
\newblock In \emph{Proceedings of the IEEE/CVF Conference on Computer Vision
  and Pattern Recognition}, pp.\  6463--6474, 2024.

\bibitem[Zhang et~al.(2023{\natexlab{a}})Zhang, Wu, Liu, Zhao, Ran, Gu, Gao,
  and Shou]{show1}
David~Junhao Zhang, Jay~Zhangjie Wu, Jia-Wei Liu, Rui Zhao, Lingmin Ran, Yuchao
  Gu, Difei Gao, and Mike~Zheng Shou.
\newblock Show-1: Marrying pixel and latent diffusion models for text-to-video
  generation.
\newblock \emph{arXiv preprint arXiv:2309.15818}, 2023{\natexlab{a}}.

\bibitem[Zhang et~al.(2023{\natexlab{b}})Zhang, Rao, and Agrawala]{controlnet}
Lvmin Zhang, Anyi Rao, and Maneesh Agrawala.
\newblock Adding conditional control to text-to-image diffusion models.
\newblock In \emph{Proceedings of the IEEE/CVF International Conference on
  Computer Vision}, pp.\  3836--3847, 2023{\natexlab{b}}.

\bibitem[Zhang et~al.(2023{\natexlab{c}})Zhang, Wang, Zhang, Zhao, Yuan, Qin,
  Wang, Zhao, and Zhou]{i2vgen_xl}
Shiwei Zhang, Jiayu Wang, Yingya Zhang, Kang Zhao, Hangjie Yuan, Zhiwu Qin,
  Xiang Wang, Deli Zhao, and Jingren Zhou.
\newblock I2vgen-xl: High-quality image-to-video synthesis via cascaded
  diffusion models.
\newblock \emph{arXiv preprint arXiv:2311.04145}, 2023{\natexlab{c}}.

\bibitem[Zhao et~al.(2023)Zhao, Gu, Wu, Zhang, Liu, Wu, Keppo, and
  Shou]{motionDirector}
Rui Zhao, Yuchao Gu, Jay~Zhangjie Wu, David~Junhao Zhang, Jiawei Liu, Weijia
  Wu, Jussi Keppo, and Mike~Zheng Shou.
\newblock Motiondirector: Motion customization of text-to-video diffusion
  models.
\newblock \emph{arXiv preprint arXiv:2310.08465}, 2023.

\bibitem[Zhao et~al.(2024)Zhao, Yuan, Wei, Zhang, Gu, Ran, Wang, Wu, Zhang,
  Zhang, et~al.]{zhao2024evolvedirector}
Rui Zhao, Hangjie Yuan, Yujie Wei, Shiwei Zhang, Yuchao Gu, Lingmin Ran, Xiang
  Wang, Zhangjie Wu, Junhao Zhang, Yingya Zhang, et~al.
\newblock Evolvedirector: Approaching advanced text-to-image generation with
  large vision-language models.
\newblock \emph{arXiv preprint arXiv:2410.07133}, 2024.

\bibitem[Zhou et~al.(2022)Zhou, Wang, Yan, Lv, Zhu, and Feng]{magicVideo}
Daquan Zhou, Weimin Wang, Hanshu Yan, Weiwei Lv, Yizhe Zhu, and Jiashi Feng.
\newblock Magicvideo: Efficient video generation with latent diffusion models.
\newblock \emph{arXiv preprint arXiv:2211.11018}, 2022.

\end{thebibliography}
\end{document}